\documentclass{article}

\usepackage[utf8]{inputenc} 
\usepackage[T1]{fontenc}    
\usepackage{caption}
\usepackage{xltabular} 
\usepackage{booktabs}  
\usepackage{array}     
\usepackage{url}

\usepackage{lmodern}
\usepackage{textcomp}
\usepackage{setspace}
\usepackage{pdfpages}

\usepackage{amsmath}
\usepackage{graphicx}
\usepackage{float}

\usepackage{tcolorbox}    
\tcbuselibrary{skins}     

\usepackage{sectsty}

\definecolor{edisongray}{HTML}{f0f0f0}  

\usepackage[left=0.9in, right=0.9in, top=0.85in, bottom=0.9in]{geometry}

\newenvironment{headerbox}{
  \begin{tcolorbox}[
    colback=edisongray,      
    colframe=edisongray,     
    frame hidden,           
    arc=4mm,                
    boxrule=0pt,            
    boxsep=0pt,             
    left=15pt, right=15pt, top=10pt, bottom=15pt,
    nobeforeafter
  ]
}{
  \end{tcolorbox}
}

\usepackage{hyperref}

\hypersetup{
  colorlinks=true,
  linkcolor=black,  
  citecolor=black,  
  urlcolor=blue     
}
\sectionfont{\sffamily}

\begin{document}

\begin{center}
\begin{headerbox}
  \vspace*{1em} 
    {\sffamily \textbf{\huge Kosmos: An AI Scientist for Autonomous\\[7pt] Discovery}}
      
    \vspace{1em}
    {Ludovico Mitchener$^{*,1,\dagger}$, Angela Yiu$^{*,1}$, Benjamin Chang$^{*,1,2}$, Mathieu Bourdenx$^{3,4,5}$, Tyler Nadolski$^{1}$, Arvis Sulovari$^{1}$, Eric C. Landsness$^{5,6}$, Dániel L. Barabási$^{7,8}$, Siddharth Narayanan$^{1}$, Nicky Evans$^{9}$, Shriya Reddy$^{10}$, Martha Foiani$^{3,4}$, Aizad Kamal$^{6}$, Leah P. Shriver$^{11,12,13}$, Fang Cao$^{10}$, Asmamaw T. Wassie$^{1}$, Jon M. Laurent$^{1}$, Edwin Melville-Green$^{1}$, Mayk Caldas$^{1}$, Albert Bou$^{1}$, Kaleigh F. Roberts$^{14}$, Sladjana Zagorac$^{15}$, Timothy C. Orr$^{6}$, Miranda E. Orr$^{6,16}$, Kevin J. Zwezdaryk$^{17,18,19}$, Ali E. Ghareeb$^{1}$, Laurie McCoy$^{1}$, Bruna Gomes$^{10}$, Euan A. Ashley$^{10}$, Karen E. Duff$^{3,4,5}$, Tonio Buonassisi$^{9,20}$, Tom Rainforth$^{2}$, Randall J. Bateman$^{5,6}$, Michael Skarlinski$^{1}$, Samuel G. Rodriques$^{1,7,\ddagger}$, Michaela M. Hinks$^{1,\dagger}$, Andrew D. White$^{1,7,\ddagger}$}
    \vspace{1em}
    
    \sffamily \textbf{\large Abstract}

    \vspace{0.2em}
    
    Data-driven scientific discovery requires iterative cycles of literature search, hypothesis generation, and data analysis. Substantial progress has been made towards AI agents that can automate scientific research, but all such agents remain limited in the number of actions they can take before losing coherence, thus limiting the depth of their findings. Here we present Kosmos, an AI scientist that automates data-driven discovery. Given an open-ended objective and a dataset, Kosmos runs for up to 12 hours performing cycles of parallel data analysis, literature search, and hypothesis generation before synthesizing discoveries into scientific reports. Unlike prior systems, Kosmos uses a structured world model to share information between a data analysis agent and a literature search agent. The world model enables Kosmos to coherently pursue the specified objective over 200 agent rollouts, collectively executing an average of 42,000 lines of code and reading 1,500 papers per run. Kosmos cites all statements in its reports with code or primary literature, ensuring its reasoning is traceable. Independent scientists found 79.4\% of statements in Kosmos reports to be accurate, and collaborators reported that a single 20-cycle Kosmos run performed the equivalent of 6 months of their own research time on average. Furthermore, collaborators reported that the number of valuable scientific findings generated scales linearly with Kosmos cycles (tested up to 20 cycles). We highlight seven discoveries made by Kosmos that span metabolomics, materials science, neuroscience, and statistical genetics. Three discoveries independently reproduce findings from preprinted or unpublished manuscripts that were not accessed by Kosmos at runtime, while four make novel contributions to the scientific literature.
    \vspace{1em} 
    
    {\footnotesize
    \raggedright
    $^{1}$Edison Scientific Inc., San Francisco, CA, USA \\
    $^{2}$University of Oxford, Oxford, UK \\
    $^{3}$UK Dementia Research Institute at University College London, London, UK \\
    $^{4}$Queen Square Institute of Neurology, University College London, London, UK \\
    $^{5}$Consortium for Biomedical Research and Artificial Intelligence in Neurodegeneration (C-BRAIN) \\
    $^{6}$Department of Neurology, Washington University School of Medicine, St. Louis, MO, USA \\
    $^{7}$FutureHouse Inc., San Francisco, CA, USA \\
    $^{8}$Eric and Wendy Schmidt Center, Broad Institute of MIT and Harvard, Cambridge, MA, USA \\
    $^{9}$Research Laboratory of Electronics, Massachusetts Institute of Technology, Cambridge, MA, USA \\
    $^{10}$Division of Cardiovascular Medicine, Stanford University, CA, USA \\
    $^{11}$Department of Chemistry, Washington University, St. Louis, MO, USA \\
    $^{12}$Center for Mass Spectrometry and Metabolic Tracing, Washington University, St. Louis, MO, USA \\
    $^{13}$Division of Nutritional Science and Obesity Medicine, Washington University, St. Louis, MO, USA \\
    $^{14}$Department of Pathology and Immunology, Washington University, St. Louis, MO, USA \\
    $^{15}$Growth Factors, Nutrients and Cancer Group, Molecular Oncology Programme, Spanish National Cancer Research Center, Madrid, Spain \\
    $^{16}$St. Louis VA Medical Center, St Louis, MO, USA \\
    $^{17}$Department of Microbiology and Immunology, Tulane University School of Medicine, New Orleans, LA, USA \\
    $^{18}$Tulane Center for Aging, Tulane University School of Medicine, New Orleans, LA, USA \\
    $^{19}$Tulane Brain Institute, Tulane University School of Medicine, New Orleans, LA, USA \\
    $^{20}$Centre d'Electronique et de Microtechnique (CSEM), Neuchâtel, Switzerland \\
    \smallskip
    {$^{*}$These authors contributed equally.} \\
    {$^{\ddagger}$These authors jointly supervise work at Edison.} \\
    {$^{\dagger}$These authors jointly supervised this work.} \\
    Correspondence to \texttt{\{andrew,sam\}$@$edisonscientific.com}\\
    }
    \vspace{-1.5cm}\hfill\raisebox{0cm}{\includegraphics[width=1.3cm]{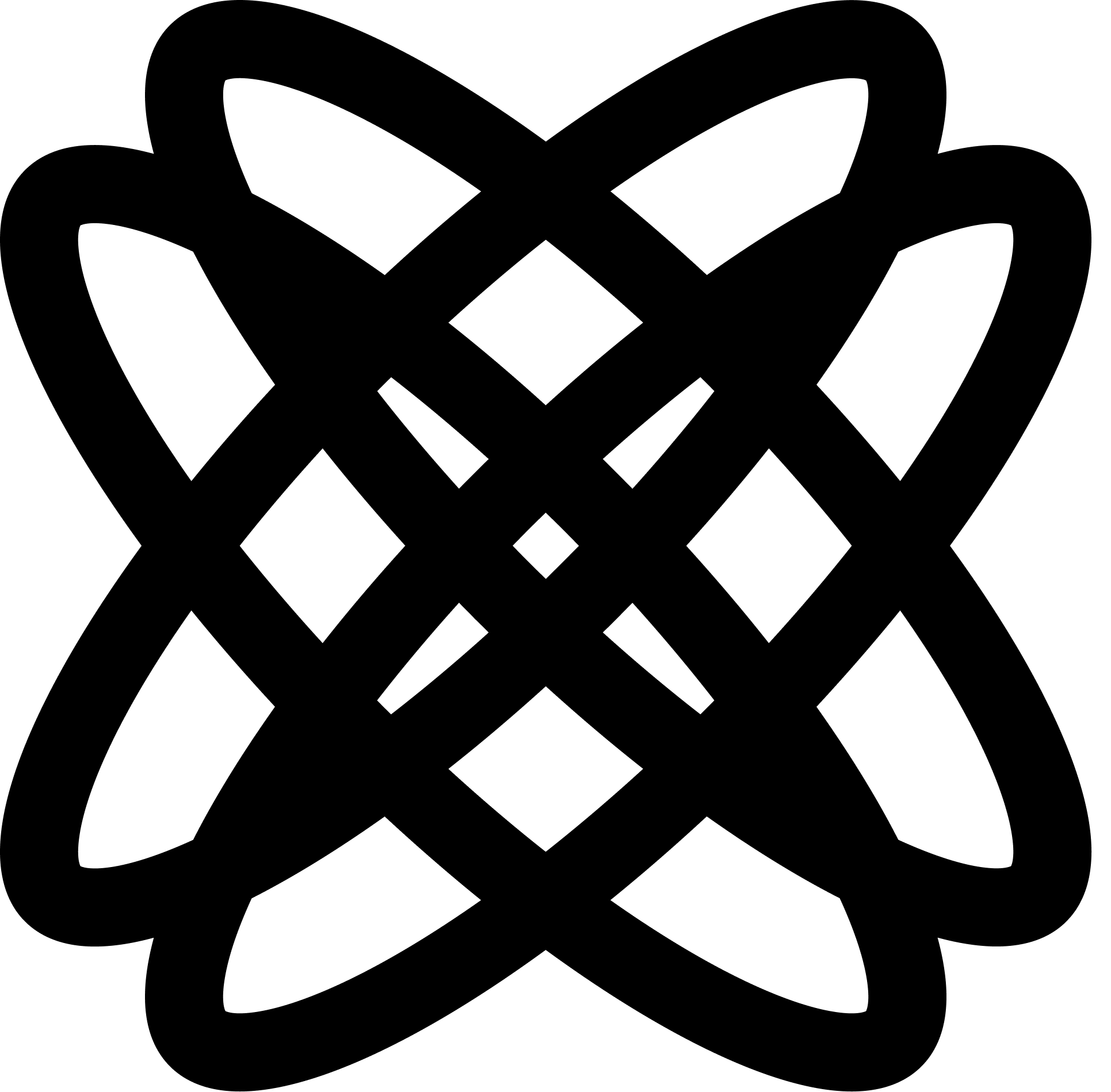}}\hspace{0cm}
    
    
\end{headerbox}
\end{center} 

\normalsize
\section{Introduction}

Data-driven discovery consists of iterative steps of literature search, hypothesis generation, and data analysis to draw new scientific conclusions from datasets. Due to their proficiency in programming and interdisciplinary reasoning, large language model (LLM) agents have the potential to automate data-driven discovery across domains.

Robin~\cite{ghareeb_robin_2025}, a system we previously reported, performs automated cycles of literature search and data analysis to propose evidence-based hypotheses, but has limited context sharing between its agents and is primarily tailored for therapeutics development. Sakana's AI Scientist~\cite{lu_ai_2024} autonomously forms hypotheses, iteratively conducts computational experiments, and writes and reviews manuscripts about its results, but remains limited to machine learning research. Google’s AI co-scientist~\cite{gottweis_towards_2025} conducts iterative cycles of reasoning to generate scientific hypotheses, but does not perform or analyze experiments. The Virtual Lab~\cite{swanson_virtual_2025} successfully designed novel nanobodies that neutralize SARS-CoV-2, and may be extensible to other domains, but lacks exploratory data analysis capabilities.

\begin{figure}[t!p]
\includegraphics[width=\columnwidth]{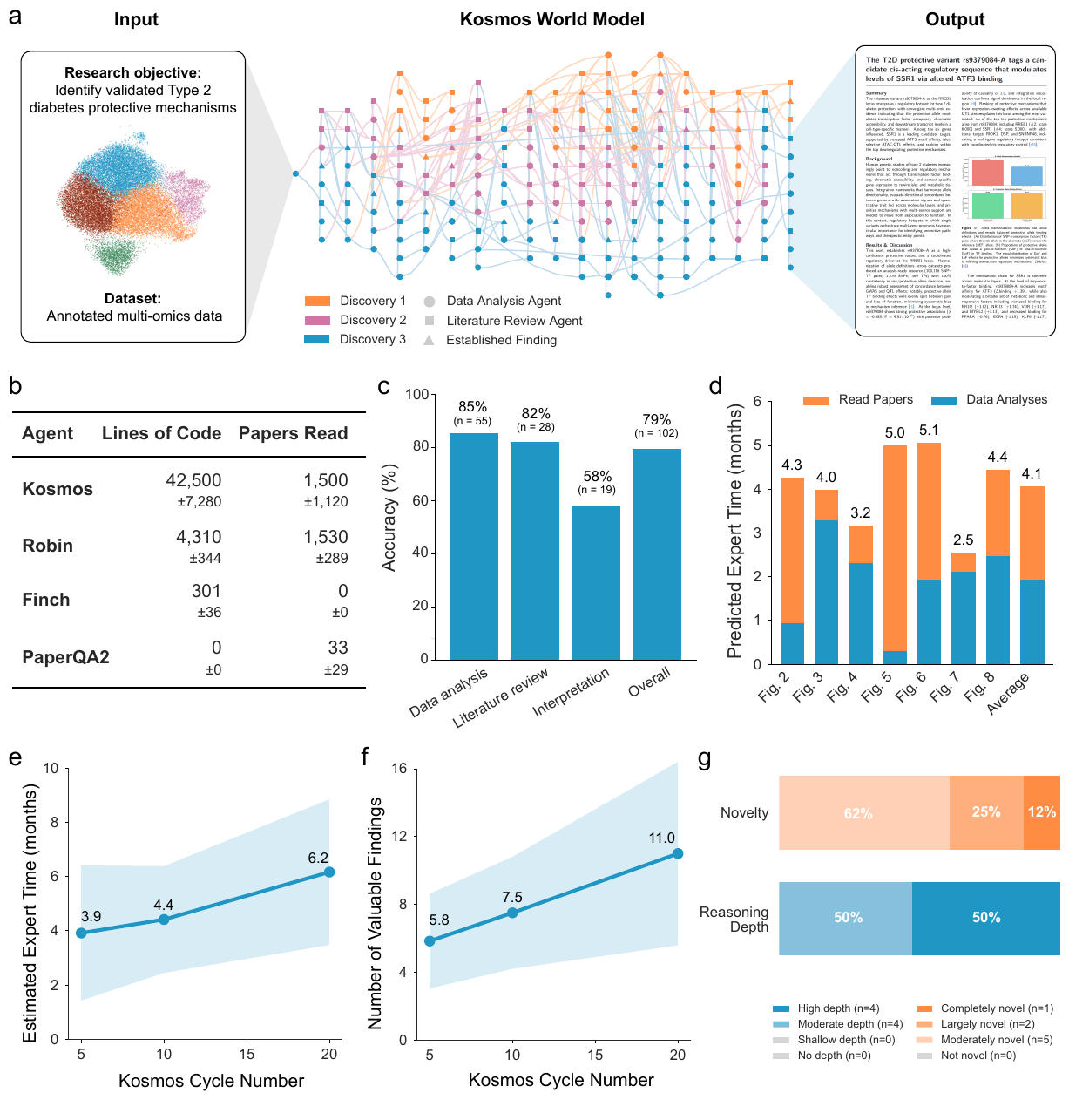}
\caption{\textbf{Kosmos workflow and performance.} \textbf{a)} Overall workflow for Kosmos. (left) Kosmos is provided with an initial dataset and broad research objective specified by a scientist. (middle) The Kosmos world model coordinates data analysis and literature search agents to identify key discoveries. (right) Each discovery is presented in a scientific report. \textbf{b)} Statistics for lines of code written and papers read for an average run across scientific agents~\cite{mitchener_bixbench_2025, skarlinski_language_2024}. \textbf{c)} Accuracy of Kosmos across data analysis, literature review, and interpretation statement types, as evaluated by expert scientists from 102 statements across three representative reports. \textbf{d)} Predicted equivalent expert time for the seven Kosmos runs described in this report, assuming it takes an expert 15 minutes to read a full paper and 2 hours to complete a Jupyter notebook of data analysis at 174 hours of work per month. \textbf{e)} Equivalent expert time was estimated by leading academic groups about the time required to achieve findings for Kosmos at cycles 5, 10, and 20, showing scaling in expert-equivalent research time with Kosmos runtime. Shaded region denotes $\pm$1 SD. \textbf{f)} The number of valuable findings generated at different steps of a Kosmos run scales with Kosmos runtime, as estimated by leading academic groups. Shaded region denotes $\pm$1 SD. \textbf{g)} Academic groups' evaluations of valuable findings from Kosmos cycle 20 indicate moderate to complete novelty (top) and high to moderate reasoning depth (bottom).}
\label{fig:fig1}
\end{figure}

Here, we present Kosmos, an AI scientist that automates data-driven discovery across a wide range of scientific disciplines. Given an open-ended objective and a dataset, Kosmos performs iterative cycles of parallel data analysis, literature search, and hypothesis generation, and summarizes its discoveries in scientific reports (\hyperref[fig:fig1]{Figure~1a}). At each cycle, Kosmos launches several parallel instances of two general-purpose Edison Scientific agents, a data analysis agent~\cite{mitchener_bixbench_2025} and a literature search agent~\cite{skarlinski_language_2024}, with each instance assigned to a specific task that is aligned with the end objective. Kosmos shares and synthesizes information among these agents by continuously updating a structured world model, which enables Kosmos to execute an average of 42,000 lines of code across 166 data analysis agent rollouts and read 1,500 full-length scientific papers across 36 literature review agent rollouts per run (\hyperref[fig:fig1]{Figure~1b}). This is a 9.8x increase in code generation compared to Robin. Consolidating information in the world model further allows every claim in a Kosmos scientific report to be directly linked to the data analysis or source from which it originated, ensuring that Kosmos’ reasoning is traceable. 

We report seven discoveries made by Kosmos: three discoveries made by Kosmos reproduce findings from preprinted or unpublished manuscripts, while the remaining four make novel contributions to the scientific literature. Each discovery is derived from a unique data type and field and is corroborated by independent analysis from a domain expert. Among the examples below, Kosmos identified a novel, clinically relevant mechanism of neuronal aging, and generated novel statistical evidence that high circulating levels of superoxide dismutase 2 (SOD2) may causally reduce myocardial fibrosis in humans. Together, these discoveries illustrate a system that can autonomously reproduce, refine, and generate data-driven discoveries with the rigor and transparency essential for advancing scientific understanding.

\section{Results}

\subsection{Kosmos system and architecture}

The core advancement in Kosmos is the use of a structured world model to manage the output of a large number of agents running in parallel. Kosmos is initiated with a research objective and a dataset, which are specified by a scientist. Kosmos attempts to complete the research objective by using LLMs, data analysis agents, literature search agents, and the world model to perform iterative discovery cycles. In each cycle, Kosmos executes up to ten literature search and analysis tasks, and subsequently updates the world model with summaries of the task outputs. Kosmos then queries the world model to propose literature search and data analysis tasks to be completed in the next cycle. This context management strategy allows Kosmos to explore many different research avenues simultaneously, and run for eight times as many iterations than existing systems~\cite{ghareeb_robin_2025, lu_ai_2024, zou_agente_2025}. Once Kosmos believes it has completed the research objective, it synthesizes key discoveries into three or four scientific reports. Each statement and figure in the report cites either a publication found by the literature search agent or a Jupyter notebook created by the data analysis agent. 

To assess the overall accuracy of Kosmos reports, we first extracted 102 statements from three representative reports and determined whether the statements originated from the scientific literature, a data analysis, or an interpretation between the two. We then asked expert scientist evaluators to classify the accuracy of each statement as “Supported” or “Refuted”. The scientist evaluators were instructed to classify the statement as “Supported” if they could reproduce the statement with their own analysis or find support for the statement in the literature, or “Refuted” if their analyses or literature search produced a different result (see \hyperref[sec:methods]{Methods}). The original code or cited paper(s) supporting the statements were not made available to the evaluating scientist during this process. Overall, 79.4\% of the statements in the report were accurate, with differing results by type: 85.5\% of the data analysis-based statements were reproducible, 82.1\% of literature review-based statements were validated with primary sources, and 57.9\% of synthesis statements were accurate (\hyperref[fig:fig1]{Figure~1c}).

We then evaluated the time it would take for a human scientist to complete the work that Kosmos performs in an individual run. We first calculated this value by tallying the number of data analysis and papers included in a given Kosmos run and estimating the time it would take a human researcher to complete the same number of tasks. We estimated each Kosmos run performs approximately 4.1 expert-months of research (n=6, $\sigma$=0.85; \hyperref[fig:fig1]{Figure~1d}), assuming an expert scientist takes 15 minutes to read a full paper and 2 hours to complete a data analysis task using a Jupyter notebook~\cite{kwa_measuring_2025} at 174 hours of work per month.

To obtain an orthogonal estimate of the amount of work done by Kosmos, we collaborated with leading academic groups to evaluate Kosmos at cycles 5, 10, and 20 of its run. These groups estimated that the findings from a 20-cycle Kosmos run would have taken them 6.14 months of research to complete (n=7, $\sigma$=2.49), much higher than our estimated time savings given their runs (\hyperref[fig:fig1]{Figure~1e}). Furthermore, they report that expert-equivalent research time scales with Kosmos runtime, roughly doubling from cycle 5 to cycle 20 (\hyperref[fig:fig1]{Figure~1e}). Similarly, when asked about the number of valuable findings that Kosmos generated across the run, experts report that the number of valuable findings scales with the amount of cycles in the Kosmos run (\hyperref[fig:fig1]{Figure~1f}). Lastly, when asked about Kosmos' reasoning depth and novelty, expert scientists report that valuable findings from cycle 20 demonstrate high to moderate reasoning depth and moderate to complete novelty (\hyperref[fig:fig1]{Figure~1g}). Together, these results suggest scaling between computational investment and scientific output.

In the following sections, we present seven studies illustrating discoveries made by Kosmos during these collaborations. We group these Kosmos discoveries into the following categories:
\begin{enumerate}
\item Two runs that reproduce existing discoveries that were either unpublished, or published after the cutoff of the relevant language models and not accessed by Kosmos at runtime.
\item One run that reproduces a published finding not accessed by Kosmos at runtime using independent reasoning.
\item Two runs that establish additional, novel support for existing discoveries.
\item One run that independently develops a new analytical method.
\item One run that makes a novel, clinically-relevant discovery not previously identified by human researchers.
\end{enumerate}

The prompts and datasets given to Kosmos for each of these discoveries can be found in \hyperref[sec:supp_inputs]{Supplementary Information 1}. The Kosmos reports describing these discoveries can be viewed at the links in \hyperref[tab:kosmos_reports_links]{Supplementary Table 1}. In the main figures of this report, any plots or results generated by Kosmos are highlighted in blue “Kosmos” sections. Plots generated by human scientists to validate Kosmos' results are highlighted in orange ``Human Validation'' sections. The Jupyter notebooks Kosmos wrote to generate these plots or results are linked in the figure captions. We improved the legibility of these plots for publication, but did not change the plot content.

\subsection{Kosmos replicates human findings from different fields}

\subsubsection{Discovery 1: Nucleotide metabolism as the dominant pathway altered under hypothermic conditions in brain}

\begin{figure}[t!p]
\includegraphics[width=\columnwidth]{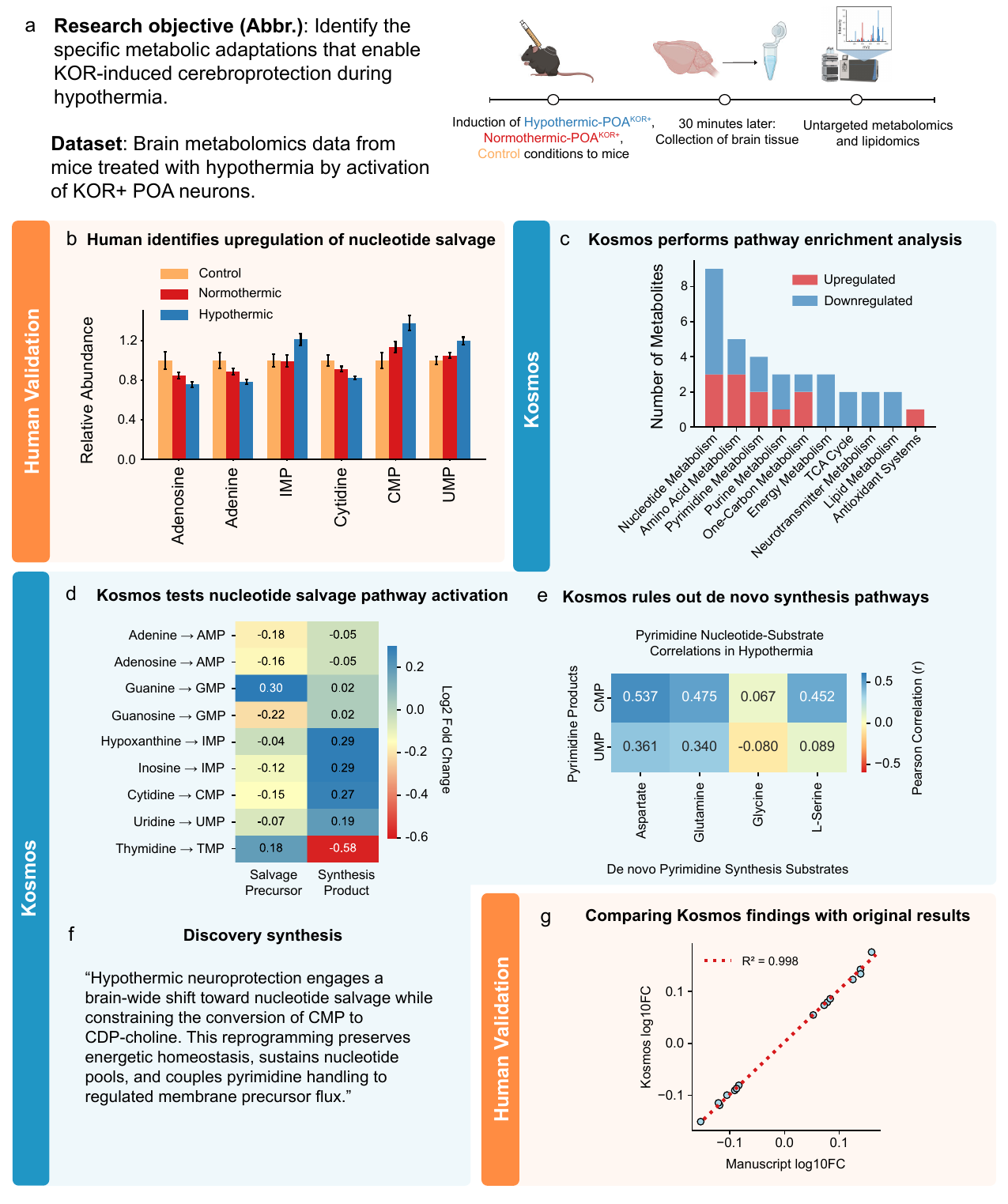}
\caption{\textbf{Kosmos reproduces unpublished discovery in neuroprotection metabolomics.} \textbf{a)} Research objective and dataset description provided to Kosmos and experimental design for untargeted metabolomic profiling of mouse brains. \textbf{b)} Barplot showing up-regulation of purine and pyrimidine salvage products after activation of POA$^{\text{KOR}+}$ neurons. Relative abundance is expressed as mean $\pm$ SEM. \textbf{c)} Top enriched pathways among metabolites that differed significantly between hypothermic and normothermic conditions \href{https://platform.edisonscientific.com/trajectories/cab1753f-30e6-4f4c-b690-ec235ff0ec6c}{[Trajectory r2]}. \textbf{d)} Heatmap showing average log$_2$ fold-change for nucleotide-salvage precursor-product pairs. Blue = increase in hypothermia relative to normothermia, and red = decrease \href{https://platform.edisonscientific.com/trajectories/c8c43d98-a499-49e4-ae2e-b75c5179c7a5}{[Trajectory r11]}. \textbf{e)} Heatmap showing Pearson correlations between pyrimidine-synthesis substrates and products. Blue = positive correlation, red = negative correlation \href{https://platform.edisonscientific.com/trajectories/3385baaa-8c1a-4b0e-bf68-69d224a0a60b}{[Trajectory r15]}. \textbf{f)} Kosmos discovery report excerpt. \textbf{g)} Scatter plot comparing Kosmos vs. original analysis: average log$_{10}$ fold change for top 15 metabolites (hypothermic vs. control). \textbf{a} and \textbf{b} reproduced with permission from~\cite{kamal_preoptic_2025}.}
\label{fig:fig2}
\end{figure}

We tested whether Kosmos could reproduce an unpublished discovery using metabolomics data (\hyperref[fig:fig2]{Figure~2a}). The data used in this run was originally collected to help identify the metabolic mechanisms underlying cooling-induced neuroprotection. While the Kosmos run was performed on the unpublished dataset, this work has since been preprinted~\cite{kamal_preoptic_2025}. Specifically, the original experiment investigated whether activating a specific brain circuit that regulates body temperature in mice could induce a controlled, torpor-like state that protects the brain from injury. Using chemogenetic tools, Kamal \textit{et al.} selectively activated kappa opioid receptor–expressing (KOR$^{+}$) neurons in the medial preoptic area (POA), a key center for thermoregulation~\cite{kamal_preoptic_2025}.

Three groups of mice were studied after this neural activation: a control group (N = 6), a group allowed to cool naturally (hypothermic-POA\textsuperscript{KOR+}; \textasciitilde22~$^{\circ}$C, N = 6), and a group kept warm to prevent cooling (normothermic-POA\textsuperscript{KOR+}; 36~$^{\circ}$C, N = 5). Only the hypothermic group showed significantly smaller brain injuries and preserved motor performance compared to controls. In contrast, normothermic activation provided no behavioral benefit, indicating that the cerebroprotective effect depends on the development of hypothermia.

To understand the underlying biological mechanisms accompanying this protective state, Kamal \textit{et al.} performed untargeted metabolomics on whole brains collected 30 minutes after activation, capturing early circuit-driven metabolic changes independent of ischemic injury (\hyperref[fig:fig2]{Figure~2a}). The analysis revealed coordinated shifts in energy-related metabolites, particularly those involved in nucleotide, phospholipid, and sphingolipid metabolism (\hyperref[fig:fig2]{Figure~2b}), suggesting that brain cooling triggers a rapid, energy-conserving metabolic program. In particular, Kamal \textit{et al.} noted in their Discussion that ``enhanced purine and pyrimidine salvage could support energy conservation and macromolecular repair by maintaining Adenosine Triphosphate (ATP) levels and nucleotide pools''~\cite{kamal_preoptic_2025}.

Kosmos was provided with the same liquid chromatography--mass spectrometry (LC--MS) dataset used in the original study, consisting of polar metabolite intensities from mouse brains in three conditions (control, normothermic-POA\textsuperscript{KOR+}, and hypothermic-POA\textsuperscript{KOR+}). The research objective supplied to Kosmos was to identify which metabolic changes might explain why the hypothermic state is protective (\hyperref[fig:fig2]{Figure~2a}).

During its exploratory analysis, Kosmos verified balanced group design and acceptable variance among biological replicates, then determined that the wide dynamic range required a $\log_{10}$ transformation to stabilize variance before statistical testing. Differential abundance analysis identified metabolites differing most strongly under hypothermic conditions, including several nucleotides and amino-acid derivatives. Kosmos performed pathway enrichment analysis, which revealed nucleotide metabolism as the most affected pathway (\hyperref[fig:fig2]{Figure~2c}). These results were then followed by iterative hypothesis exploration proposing that nucleotide turnover could underlie the hypothermic state’s protective effects (\hyperref[fig:fig2]{Figure~2d,e}).

Kosmos concentrated on nucleotide metabolites and detected an inversion pattern: precursor bases and nucleosides decreased, while phosphorylated nucleotide products increased, consistent with activation of nucleotide-salvage pathways (\hyperref[fig:fig2]{Figure~2d}). In parallel, the literature-review agent conducted a literature search and it confirmed that nucleotide salvage is a conserved, energy-efficient strategy during hypoxia and hypothermia, reinforcing Kosmos’s emerging hypothesis~\cite{ohler_pyrimidine_2019, thauerer_purine_2012, shen_inosine_2005, benowitz_inosine_1999}. Kosmos then tested competing explanations by correlating nucleotide products with de novo synthesis substrates. Finding no strong relationships, it ruled out substrate-driven increases and concluded that the data most likely reflected upregulated salvage activity rather than new synthesis (\hyperref[fig:fig2]{Figure~2e}). Finally, Kosmos synthesized its mechanistic interpretation of the engagement of a nucleotide-salvage pathway (\hyperref[fig:fig2]{Figure~2f}).

Kosmos’s analysis closely reproduced the reasoning, results, and conclusions reached by the original study~\cite{kamal_preoptic_2025}. Both analyses identified nucleotide metabolism as the dominant pathway altered under hypothermic conditions, with nearly identical effect sizes and directional changes across key metabolites. Kosmos independently highlighted increases in inosine monophosphate (IMP), cytidine monophosphate (CMP), and uridine monophosphate (UMP) and decreases in adenine, cytidine, and Guanosine diphosphate (GDP)-mannose, reproducing the same precursor-to-product inversion pattern described by the authors as evidence of activated nucleotide salvage. Quantitatively, nine of the top fifteen metabolites identified by Kosmos overlapped exactly with those reported in the original analysis, and the $\log_{10}$ fold-change estimates of the top 15 metabolites is nearly indistinguishable ($R^2 = 0.998$, \hyperref[fig:fig2]{Figure~2g}).

Kosmos also extended the human analysis by attempting to rule out contributions from de novo nucleotide-synthesis pathways. While this step demonstrated initiative in its reasoning, it also highlighted a key limitation: the dataset contained only steady-state metabolite levels rather than the time-resolved flux data required to assess pathway activity, so its conclusion regarding de novo synthesis cannot be considered definitive. These results demonstrate that Kosmos not only replicated the quantitative outcomes of the human analysis but also arrived at the same mechanistic conclusion: activation of POA\textsuperscript{KOR+} neurons produces a torpor-like hypothermic state characterized by selective engagement of energy-efficient metabolic programs, notably the nucleotide-salvage pathway, which supports cellular energy balance and macromolecular maintenance under reduced thermometabolic demand.

\subsubsection{Discovery 2: Thermal annealing humidity as critical determinant for perovskite solar-cell performance}

\begin{figure}[t!p]
\includegraphics[width=\columnwidth]{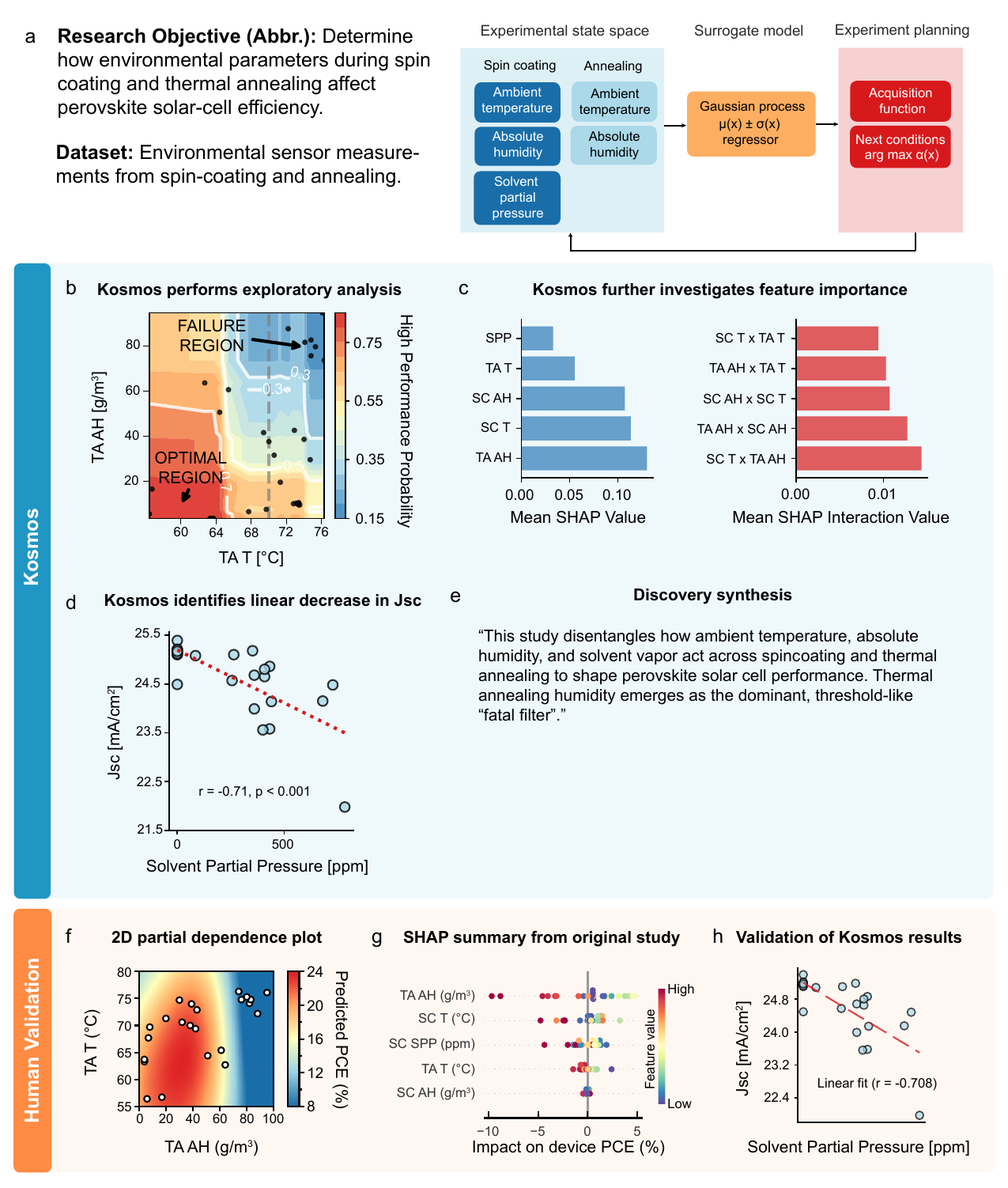}
\caption{\textbf{Kosmos reproduces perovskite solar cell fabrication findings.} \textbf{a)} Research objective and dataset supplied to Kosmos. Schematic of modeling approach from original preprint. \textbf{b)} 2D partial dependence plot showing the predicted high performance probability (color scale) as a function of thermal annealing temperature and absolute humidity, based on a random forest model. \href{https://platform.edisonscientific.com/trajectories/4d090202-3d76-49ef-9879-82748427e8dc}{[Trajectory r31]}. \textbf{c)} Mean absolute SHAP values for (left) single variables and (right) their top interaction effects on device performance \href{https://platform.edisonscientific.com/trajectories/a4df32f6-3169-4e2c-8d22-ae4c7c854bdb}{[Trajectory r44]}. SPP, solvent partial pressure of DMF; TA T, thermal annealing temperature; SC AH, spin-coating absolute humidity; SC T, spin-coating temperature; TA AH, thermal annealing absolute humidity. \textbf{d)} Short-circuit current density (J$_\text{SC}$) decreases linearly with increasing DMF SPP for non-failure devices (n=23) \href{https://platform.edisonscientific.com/trajectories/91a8373d-6a6a-4b6b-8acd-8ee79255aae2}{[Trajectory r81]}. \textbf{e)} Kosmos discovery report excerpt. \textbf{f)} Joint influence of temperature and absolute humidity during thermal annealing. Average PCE (colour scale) predicted by Gaussian process regressor. \textbf{g)} Influence of environmental variables on PSC PCEs, SHAP values derived from Gaussian process regressor. \textbf{h)} Scatter plot and a linear fit for J$_\text{SC}$ and DMF SPP values for a subset of devices.}
\label{fig:fig3}
\end{figure}

Having demonstrated Kosmos' ability to reproduce published findings using metabolomics data, we next explored whether Kosmos could be extended beyond biological datasets. We provided Kosmos with a material science dataset from Liu \textit{et al.}, arXiv 2025~\cite{liu_disentangling_2025} describing environmental parameters from fabrication of solar cells with an open-ended research objective (\hyperref[fig:fig3]{Figure~3a}).

Repeatable device performance of perovskite solar cells remains a major challenge, with small variations in environmental conditions during fabrication---such as temperature, humidity, and solvent partial pressure---strongly influencing how the material crystallizes and how well the final device performs. However, the combined effects of these factors are not yet fully understood. Liu \textit{et al.} introduces a systematic approach to study how individual and interacting environmental variables affect perovskite film formation and device efficiency. The researchers developed a fabrication platform that allows precise, independent control of ambient temperature, absolute humidity, and solvent partial pressure during key processing steps~\cite{liu_disentangling_2025}. Using this platform, a closed-loop Bayesian optimization strategy was applied to efficiently explore and map how these variables influence device outcomes (\hyperref[fig:fig3]{Figure~3a, right}). These findings demonstrate that integrated environmental control is essential for producing reproducible perovskite solar cells, and highlight the value of interpretable machine learning to navigate complex processing environments.

The results demonstrate that ambient temperature, absolute humidity, and solvent partial pressure during perovskite film formation strongly affect solar cell efficiency. Liu \textit{et al.} quantified and ranked the influence of each parameter, revealing both individual and coupled non-linear interactions that govern device performance. Analysis by Kosmos corroborated these findings, identifying absolute humidity during thermal annealing as the most dominant factor determining device efficiency under single-variable analysis, while also exhibiting interaction effects (\hyperref[fig:fig3]{Figure~3b}). This dominance is evident in the SHapley Additive exPlanations (SHAP) plots from Kosmos (\hyperref[fig:fig3]{Figure~3c}) and from the original study (\hyperref[fig:fig3]{Figure~3g}). While the original study emphasizes the complex and interdependent nature of environmental parameters during synthesis, highlighting the importance of environmental control for reproducible fabrication, Kosmos’ extended analysis further delineates performance regimes and quantifies their effects on specific device metrics.

Together, these insights define environmental guidelines for optimizing perovskite film growth using the experimental apparatus described in Liu \textit{et al.} Both analyses identified thermal annealing humidity as a critical determinant, acting as a “fatal filter” beyond which devices fail. The 2D partial dependence plot (\hyperref[fig:fig3]{Figure~3b}) shows this discrete failure boundary occurring above \textasciitilde60~g/m\textsuperscript{3} absolute humidity and within a critical temperature window above \textasciitilde70~$^{\circ}$C, with optimal performance at low humidity and moderate temperatures. This trend is independently confirmed in the human-made plot from the Liu \textit{et al.} preprint (\hyperref[fig:fig3]{Figure~3f}). Additionally, Kosmos revealed a linear decrease in short-circuit current density ($J_{SC}$) with rising Dimethylformamide (DMF) solvent partial pressure during spin-coating (\hyperref[fig:fig3]{Figure~3d})---a previously unreported relationship that was later independently validated by the researchers (\hyperref[fig:fig3]{Figure~3h}).

\subsubsection{Discovery 3: Log-normal connectivity distributions in neuronal networks}

\begin{figure}[t!p]
\includegraphics[width=\columnwidth]{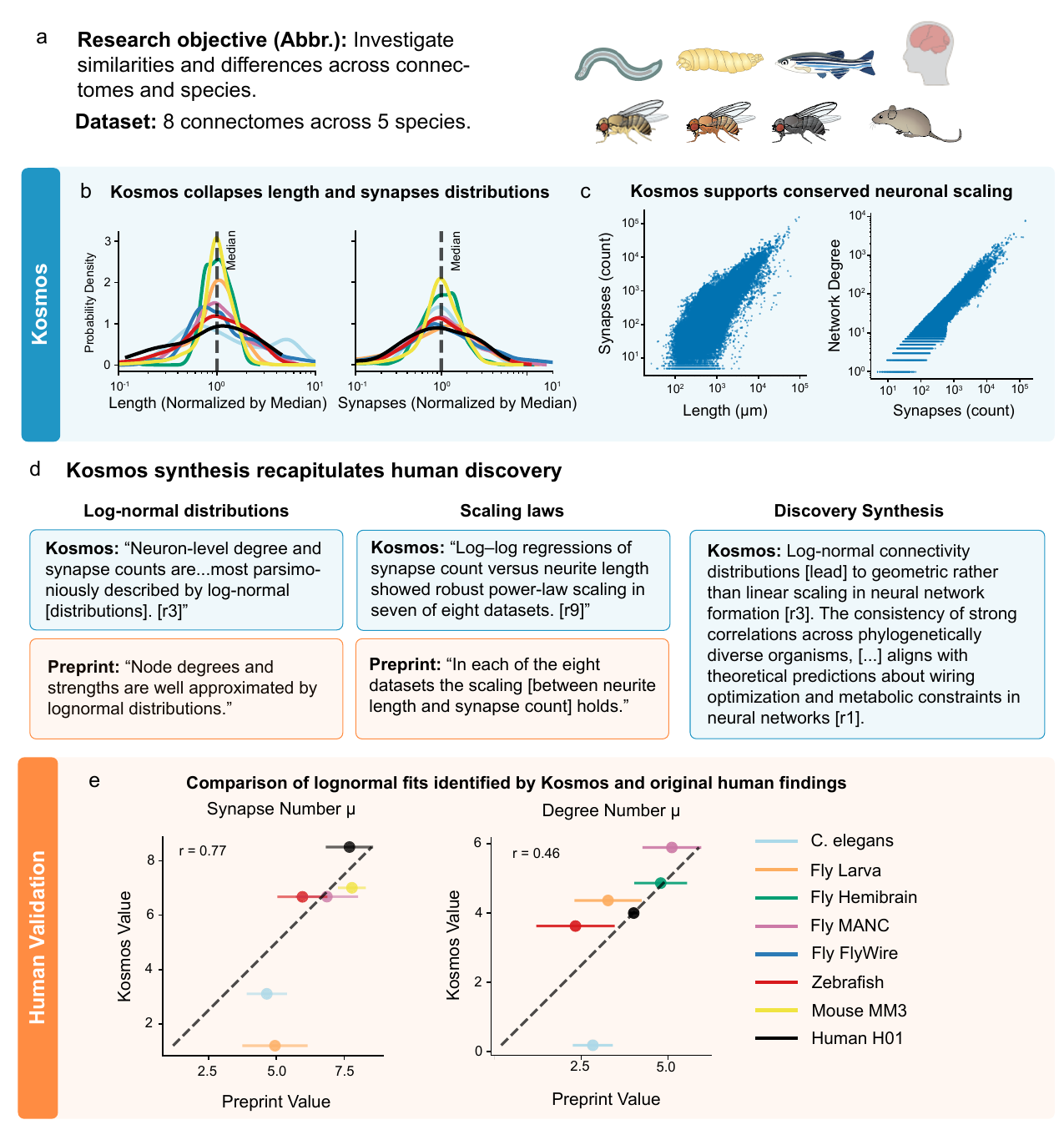}
\caption{\textbf{Kosmos reproduces findings on log-normal connectivity distributions in neural networks.} \textbf{a)} Kosmos was provided with a brief research objective and preprocessed connectomic data from eight separate reconstructions, spanning 5 species. \textbf{b)} Kosmos normalized neuronal wire length (left) and neuronal synapse count (right) \href{https://platform.edisonscientific.com/trajectories/9f4c861d-4ac4-4189-bedb-4d49374ffd44}{[Trajectory r7]}. Plots are Kosmos' visualization with colors corresponding to datasets as per legend in (e). \textbf{c)} Kosmos investigated correlations between Length, Synapses and Degree \href{https://platform.edisonscientific.com/trajectories/9935f4d8-4691-4b68-9200-03b3b2f9bfb0}{[Trajectory r1]}. Plots show correlations for the fruit fly's central nervous system (FlyWire), although Kosmos analyzed all animals. \textbf{d)} (left) Kosmos finds that despite canonical expectations of power law scaling in neuron degree, the neuron synapse number and neuron degree overall are best fit by lognormal distributions \href{https://platform.edisonscientific.com/trajectories/a1142921-5955-436a-9d0c-55467254264f}{[Trajectory r3]}, recapitulating preprint Figure 2A and 2D. (middle) Kosmos confirms \href{https://platform.edisonscientific.com/trajectories/9935f4d8-4691-4b68-9200-03b3b2f9bfb0}{[Trajectory r1]}'s proposal of a power law relationship between synapses and Length \href{https://platform.edisonscientific.com/trajectories/77cc5b88-ebd8-4d73-8a72-082a2ba161e4}{[Trajectory r9]} , mirroring Figure 3a of the preprint. (right) Beyond reproducing the key findings of the preprint, Kosmos proposes neuroscientific underpinnings of its conclusions, which correctly identify the final major result of the preprint, that multiplicative processes can govern neuronal properties, leading to the observed lognormal distributions (preprint Figure 4). \textbf{e)} The lognormal fits for synapse number (left) and degree number (right) found by Kosmos have high concordance with the values measured in the preprint. Bars around dots indicate standard deviations reported in the preprint.}
\label{fig:fig4}
\end{figure}

We next investigated whether Kosmos could independently reproduce findings from an existing preprint when given an open-ended research objective. To do this, we provided Kosmos with the same data used in Piazza \textit{et al.}, bioRxiv 2025~\cite{piazza_physical_2025}: Neuron morphology data from eight separate connectome reconstructions from five different species (\hyperref[fig:fig4]{Figure~4a}). Each connectome dataset contained detailed information about neuronal wire length, the number of connecting partners (degree), total number of synapses, and local measures of synaptic density. We asked Kosmos to investigate how these measures relate from a neuroscientific perspective in order to propose universal principles among them. Specifically, we aimed to see whether this broad prompt was sufficient to capture the major findings of the preprint: (1) degree distributions cannot be captured by the well-established random or scale-free models, rather all morphological metrics are well approximated by lognormal distributions, and (2) the metrics can be related through power law distributions, suggesting that (3) multiplicative processes drive the development of neurons, and thus the emergence of all lognormals.

In an initial analysis, Kosmos normalized wire length, degree, and synapse density distributions by dividing by its median and standard deviation, which collapsed the distributions from different species onto a single curve (\hyperref[fig:fig4]{Figure~4b}). While distribution collapse is commonly interpreted as a qualitative metric for assessing distribution similarity, Kosmos applied a conservative approach by performing pairwise Kolmogorov-Smirnov tests between all distributions, which led to an erroneous rejection of distributional similarity despite the clear visual collapse onto a single curve, which successfully mirrors Figures~2B,E,H,K of the preprint~\cite{piazza_physical_2025}.

In parallel, Kosmos investigated correlations between wire length, total synapses, and degree across all datasets (\hyperref[fig:fig4]{Figure~4c}). The analysis revealed strong positive correlations between these neuronal properties in all datasets, with the linear pattern evident in log-log plots. These results strongly supported the hypothesis of conserved neuronal property scaling across species, successfully reproducing the core findings shown in Figure~3A,G of Piazza~\cite{piazza_physical_2025}.

Given these initial data analyses, Kosmos then successfully validated the two main findings of the preprint. 

First, Kosmos examined the underlying distributions for neuronal morphology metrics. Having observed that the datasets were heavy-tailed, it fit exponential, power law and lognormal distributions to synapse number and degree, concluding that ``neuron-level degree and synapse counts are...most parsimoniously described by log-normal'' distributions (\hyperref[fig:fig4]{Figure~4d}), thereby recapitulating the results presented in Figure~2A,D of Piazza \textit{et al.}~\cite{piazza_physical_2025}.

Second, having identified patterns indicating power-law scaling between synapses, length and degree (\hyperref[fig:fig4]{Figure~4c}), Kosmos finds that ``Log-log regressions of synapse count versus neurite length showed robust power-law scaling in seven of eight datasets,'' thus capturing Figure~3A,G, as well as equation~5, of the preprint~\cite{piazza_physical_2025} (\hyperref[fig:fig4]{Figure~4d}). 

In order to validate these findings, the researchers compared the mean ($\mu$) of the lognormals fitted by Kosmos with the values reported in the preprint, and found high concordance between the two, with most estimates falling within one standard deviation of the published values (Pearson's $r=0.77$ for Synapses, $r=0.46$ for Degrees) (\hyperref[fig:fig4]{Figure~4e}). However, the powerlaw fitting package employed by Kosmos, while popular in network science applications, has known failure modes~\cite{hoorn_problems_2020} that underpin observed large deviations and missing values. For instance, the researchers chose to ignore cases where the algorithm fitted negative $\mu$ values (Hemibrain and Flywire for Synapse Number, Mouse and FlyWire for Degree), as a negative $\mu$ would nonsensically indicate that the center of the distribution is orders of magnitude lower than any observed datapoint. 

In summary, when given morphological measurements of neurons and a broad missive to investigate universal principles, Kosmos reproduced the two major quantitative results of the underlying preprint (\hyperref[fig:fig4]{Figure~4b-e}). In addition, Kosmos properly contextualized its results, first noting that the lognormal fits refuted the universal hypothesis of power law degree distributions in [r3], then capturing the third, qualitative result of the source material (preprint Figure~4), proposing that multiplicative processes in neurodevelopment can be causative of the observed distributions (\hyperref[fig:fig4]{Figure~4d}). Thus, Kosmos could not only capture expected findings, but also can connect empirical results to novel interpretations for network science and neurodevelopment. 

We note that the source preprint Piazza \textit{et al.}\ was posted on bioRxiv on February 27, 2025 and therefore accessible to the literature search agent and possibly included in the training data for the models used in the data analysis agent during this run. As a control, we reran Kosmos with models that have a cutoff date prior to the preprint, and the results reproduced both the lognormal and scaling findings. Further, in the Kosmos results reported in \hyperref[fig:fig4]{Figure~4}, we do not see evidence of Kosmos having ``memorized'' the results reported here, nor did the literature search agent access Piazza \textit{et al.}, bioRxiv 2025. Instead, we observe Kosmos following a chain of independent analysis and reasoning to reach the same conclusion as reported in Piazza \textit{et al.}

\subsection{Kosmos adds additional support to existing findings with novel methods}

\subsubsection{Discovery 4: SOD2 as a driver of myocardial fibrosis in humans}

\begin{figure}[t!p]
\includegraphics[width=\columnwidth]{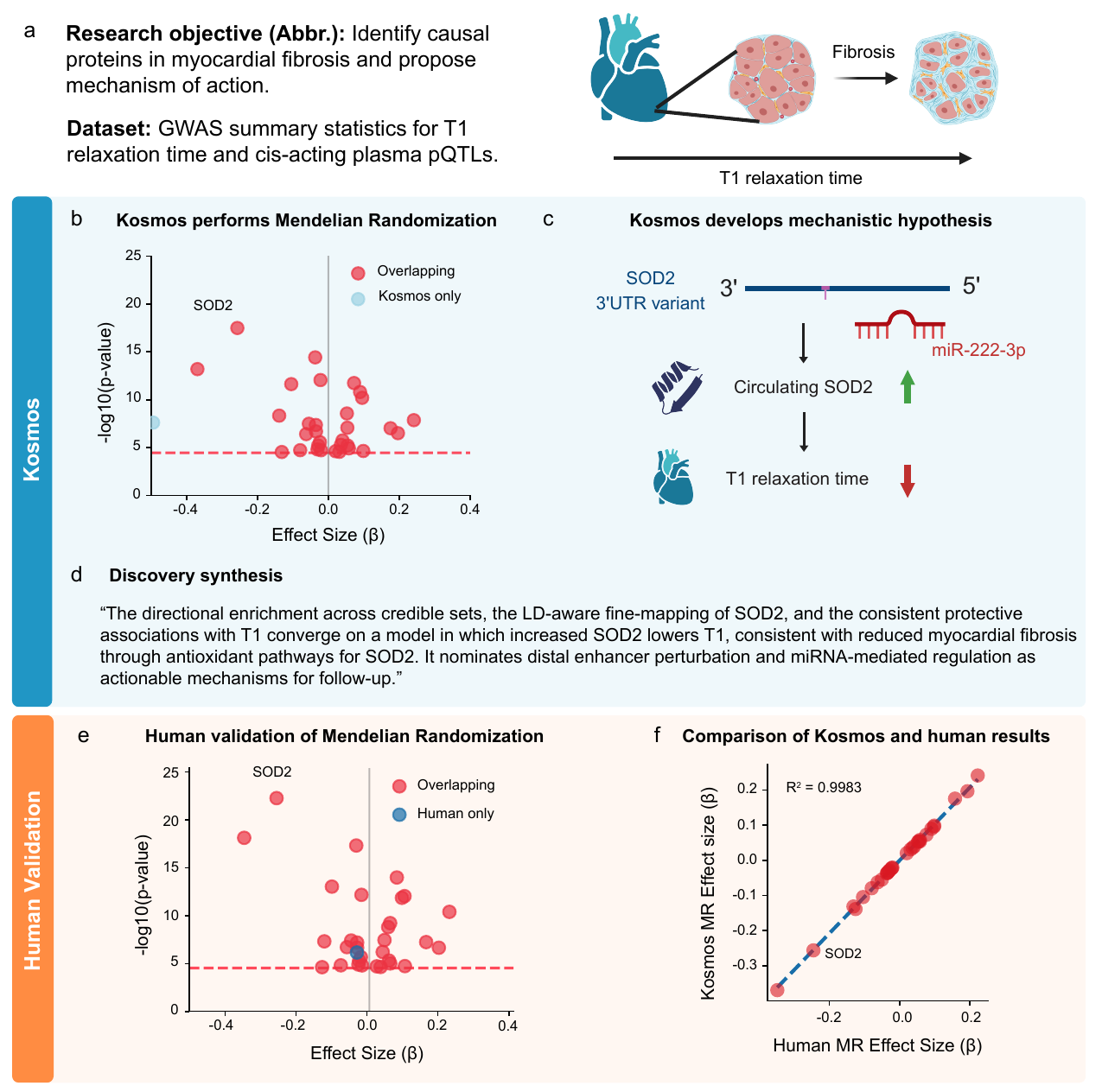}
\caption{\textbf{Kosmos identifies SOD2 as a driver of myocardial fibrosis in humans.} 
\textbf{a)} Research objective and dataset description given to Kosmos. All datasets provided are publicly available. Native T1 relaxation times measured by cardiac magnetic resonance imaging reflect myocardial tissue composition, with elevations indicating increased interstitial volume from fibrosis or edema. (Diagram Created in \underline{\url{https://BioRender.com}}) \textbf{b)} Mendelian randomization (MR) analysis performed by Kosmos. Volcano plot showing beta coefficient and associated p-values for Mendelian Randomization. Red dotted line indicates the significance threshold. Red dot shows proteins identified by Kosmos and manually, and blue indicates proteins identified by Kosmos only \href{https://platform.edisonscientific.com/trajectories/19a2137a-c953-45cb-8602-2278dadebfd7}{[Trajectory r11]}. \textbf{c)} Kosmos proposes the hypothesis that a variant in the 3$'$ UTR of SOD2 alters post-transcriptional regulation by disrupting miRNA (miR-222) binding \href{https://platform.edisonscientific.com/trajectories/1de48eea-c769-44e1-8e1c-10ef746c64f4}{[Trajectory r41]}. (Diagram Created in \underline{\url{https://BioRender.com}}) \textbf{d)} Kosmos consolidates the SOD2 results into a discovery. \textbf{e)} Volcano plot showing manual Mendelian randomization results. SOD2 was the most significant target. Red indicates overlapping proteins identified and blue indicates protein identified in human analysis only. Red dotted line indicates the significance threshold. \textbf{f)} Scatter plot showing the Mendelian Randomization effect sizes ($\beta$) as determined manually by humans and by Kosmos for the 31 overlapping proteins between the two sets.}
\label{fig:fig5}
\end{figure}

After demonstrating that Kosmos can replicate recent published and unpublished discoveries, we next sought to test if Kosmos could analyze public datasets using a recently developed analysis pipeline (Reddy \textit{et al.}, manuscript in preparation) (\hyperref[fig:fig5]{Figure~5a}).

Cardiac fibrosis is a major pathological contributor to heart failure, affecting tens of millions of patients worldwide. Cardiac fibrosis increases ventricular stiffness and impairs contractility, yet therapeutic interventions directly targeting fibrotic pathways remain elusive~\cite{ghazal_cardiac_2025}. Mendelian randomization (MR) leverages the random allocation of genetic variants at conception to infer causal relationships between exposures and outcomes. By using genetic variants as instrumental variables for protein levels, MR eliminates confounding and reverse causation. The approach is particularly powerful for identifying causal biomarkers and therapeutic targets from large-scale genomic datasets.

Reddy \textit{et al.} applied a proteome-wide MR framework to investigate causal proteins in cardiac structure and function using native T1 mapping as a quantitative proxy for myocardial fibrosis (Reddy \textit{et al.}, manuscript in preparation). Native T1 relaxation times measured by cardiac magnetic resonance imaging reflect myocardial tissue composition, with elevations indicating increased interstitial volume from fibrosis or edema. 

The same dataset, the myocardial T1 Genome Wide Association Study (GWAS) and cis-protein quantitative trait loci (pQTL) data, was provided to Kosmos with instructions to use R packages TwoSampleMR for Mendelian randomization, coloc for colocalization, and susieR for fine-mapping (\hyperref[fig:fig5]{Figure~5a}). Kosmos autonomously executed the analytical pipeline to identify protein candidates and propose mechanistic hypotheses.

The colocalization analysis failed due to cascading technical issues: many p-values stored as 0 caused Kosmos to misidentify thousands of variants as lead SNPs, column name collisions during merging led to mismatched vector lengths, and missing standard deviation and MAF information prevented proper coloc execution [r12]. Rather than attempting to repair the failed colocalization pipeline, Kosmos proceeded to SuSiE fine-mapping. This analysis identified a credible set of directionally consistent variants for superoxide dismutase 2 (SOD2) (PP.H4 = 1.00), resolving the conflicting colocalization results and providing strong evidence for pleiotropic effects. 

Both Kosmos and human MR analyses independently identified superoxide dismutase 2 (SOD2) as the primary causal protein for myocardial fibrosis, with remarkably concordant effect estimates. Kosmos reported $\beta = -0.231$ ($p = 4.23 \times 10^{-13}$) (\hyperref[fig:fig5]{Figure~5b}) while human analysis yielded $\beta = -0.258$ ($p = 1.22 \times 10^{-22}$) (\hyperref[fig:fig5]{Figure~5e}), representing nearly identical effect sizes with the same protective direction. Both approaches found near-certain colocalization evidence (KOSMOS PP.H4 = 0.9999; manual PP.H4 = 0.999999), confirming a shared causal variant between SOD2 protein levels and myocardial T1. Both Kosmos and human analysis applied Bonferroni correction to MR p-values and identified 32 proteins meeting the stringent significance threshold ($p < 3.52 \times 10^{-5}$). Of these 32 proteins, 31 were identified by both approaches, representing 96.9\% overlap. Effect sizes for these 31 shared proteins showed near-perfect correlation (Pearson $r = 0.9991$, $p = 1.42 \times 10^{-41}$, $R^2 = 0.9983$), with consistent directionality across all overlapping proteins (\hyperref[fig:fig5]{Figure~5f}).

Lastly, Kosmos proposed a potential post-transcriptional mechanism for SOD2 regulation (\hyperref[fig:fig5]{Figure~5c}), identifying rs4555948 within the SOD2 3'UTR as a credible-set variant predicted to disrupt a hsa-miR-222-3p binding site (Figure~5d). However, this site does not overlap with any predicted miR-222 binding sites based on current target prediction databases, highlighting a limitation in Kosmos’s automated annotation. Literature supports that miR-222 binds the SOD2 3'UTR and regulates SOD2 protein expression~\cite{dubois-deruy_micrornas_2017}, but without specifying the exact position of binding. Thus, while the variant itself may not directly alter a miR-222 site, the SOD2 3'UTR remains a regulatory hotspot, and post-transcriptional mechanisms cannot be excluded.

Kosmos further integrated literature implicating SOD2 in vascular redox homeostasis, proposing a plausible hemodynamic pathway linking SOD2 to cardiac remodeling through effects on blood pressure and endothelial function~\cite{arendshorst_oxidative_2024, fukai_superoxide_2011, dosunmu-ogunbi_role_2021}. This hypothesis is novel, as existing evidence for SOD2 primarily derives from its intracellular antioxidant role in murine studies~\cite{sharma_sod2_2020}. Nevertheless, since other SOD isoforms have demonstrated functional activity in circulation~\cite{kliment_extracellular_2009, jiang_superoxide_2024}, the concept of circulating SOD2 contributing to myocardial fibrosis reduction remains biologically reasonable. Downstream wet-lab validation experiments are currently underway to test these mechanistic predictions.

This analysis advances prior observational findings into causal inference. Nauffal \textit{et al.}~\cite{nauffal_genetics_2023} reported correlation, but not causation between SOD2 protein levels and myocardial fibrosis. Both analyses suggested that genetically-predicted circulating SOD2 levels may causally reduce myocardial fibrosis. This discovery of SOD2 demonstrates Kosmos’s ability to identify a target through alternative methods (statistical analysis through GWAS), and suggest a potential molecular mechanism of action.

\subsubsection{Discovery 5: Cis-regulation of SSR1 by a protective GWAS variant in Type 2 Diabetes in humans}

\begin{figure}[t!p]
\includegraphics[width=\columnwidth]{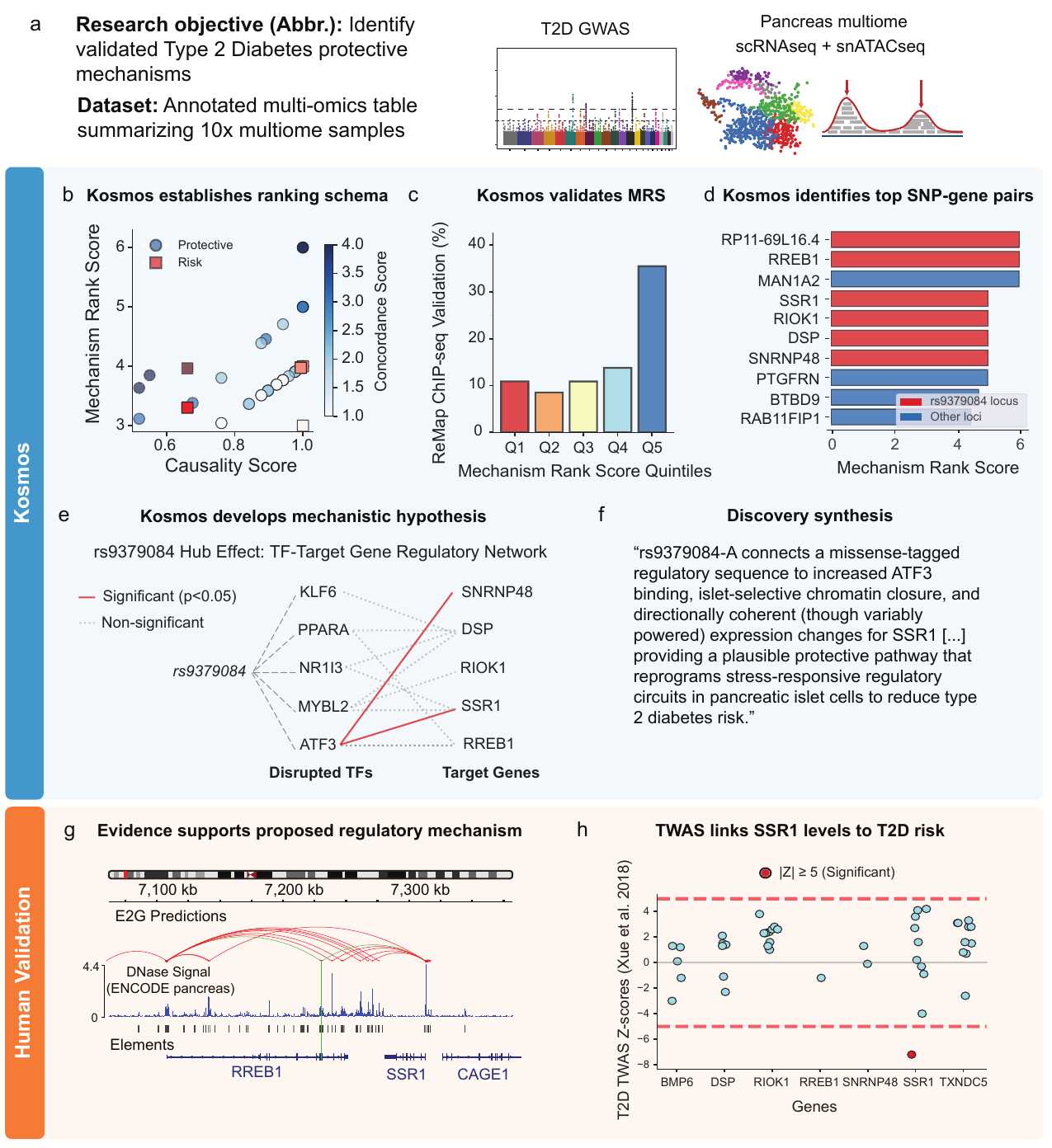}
\caption{\textbf{Kosmos identifies cis-regulation of SSR1 by a protective GWAS variant in Type 2 Diabetes.} \textbf{a)} Research objective and dataset description provided to Kosmos. (Diagram Created in \underline{\url{https://BioRender.com}}) \textbf{b)} Relationships between causality score (identical to the PIP) and Mechanism rank Score (MRS)\href{https://platform.edisonscientific.com/trajectories/45829c6f-8a4d-4ab1-bc0d-f28f04868fc4}{[Trajectory r2]}. \textbf{c)} The Q5 bin of MRS has a 3.3-fold higher ChIP-seq validation rate than Q1 ($p$-value $<$ 0.001) \href{https://platform.edisonscientific.com/trajectories/ddaae772-4079-477a-ac98-651d2ade787c}{[Trajectory r23]}. \textbf{d)} The highest MRS across the dataset is for rs9379084 locus (blue, MRS = 6), while the other red bars correspond to other protective T2D loci. \textbf{e)} Kosmos conducts an over-representation analysis: the 5 gene set from the highlighted locus in d was tested against Enrichr TF-target datasets \href{https://platform.edisonscientific.com/trajectories/7657e4fa-4622-4ca6-8c02-528fb2e1a540}{[Trajectory r53]}. The gene set was enriched for targets of ATF3 ($p=0.042$), which regulates two of the five genes (SSR1 and SNRNP48). \textbf{f)} Kosmos consolidates the analyses into a coherent mechanistic hypothesis on how the variant rs9379084 confers reduced T2D risk. \textbf{g)} An IGV screenshot of E2G data accessed here: \underline{\url{http://e2g.stanford.edu/variant/6_7231610_G_A}}. The green lines correspond to the location of the missense variant rs9379084 and the predicted cis-regulatory effect on RREB1 and SSR1. \textbf{h)} A reflection plot of the TWAS Z-scores from the TWAS hub (\underline{\url{http://twas-hub.org}}) showing that in the locus of interest, only the genetically-inferred SSR1 levels are associated with T2D risk at a transcriptome-wide significant level  ($Z = -7.2$).}
\label{fig:fig6}
\end{figure}

After demonstrating that Kosmos can follow user-provided analytical pipelines and methodological instructions, we next assessed whether Kosmos could independently prioritize and justify likely causal mechanisms from a GWAS of Type 2 Diabetes (T2D) without external guidance.

GWAS have identified thousands of common variants associated with T2D risk~\cite{vujkovic_discovery_2020}. Converting these statistical associations into mechanistic insights remains a central challenge in human genetics, requiring integration across multiple molecular data modalities. Over 1,000 genes have genetically inferred transcript levels associated with T2D risk~\cite{vujkovic_discovery_2020}, yet functional characterization of these loci has been hindered by locus complexity, linkage disequilibrium, and limited cell-type resolution.

To address this problem at scale, we provided Kosmos with a variant-level input dataset that integrated multi-omic annotations from publicly available sources, including T2D GWAS summary stats with fine-mapping posterior inclusion probabilities (PIPs)~\cite{mahajan_multi-ancestry_2022}, computational prediction for Transcription Factor Binding Sites (TFBS) overlap, chromatin accessibility (ATAC-seq)~\cite{wang_integrating_2023}, expression (eQTL)~\cite{the_gtex_consortium_gtex_2020}, and protein (pQTL)~\cite{sun_plasma_2023} data (\hyperref[fig:fig6]{Figure~6a}). Kosmos was tasked with generating mechanistic hypotheses for protective T2D variants, which were defined as biallelic variants where the alternative allele confers reduced T2D risk relative to the reference allele (\hyperref[fig:fig6]{Figure~6a}).

Without explicit instruction, Kosmos established a mechanistic ranking schema termed the Mechanistic Ranking Score (MRS), defined as: $MRS = PIP \times (1 + \text{Concordance Score} + \text{Experimental Evidence Score})$ (\hyperref[fig:fig6]{Figure~6b}) where: PIP represents the posterior inclusion probability from fine-mapping; Concordance Score quantifies the number of QTL data types (ATAC-seq, eQTL, pQTL) showing directionally consistent effects with the GWAS signal (e.g., decreased expression or chromatin accessibility for protective alleles); Experimental Evidence Score adds 1 point for predicted transcription-factor binding disruption and 1 additional point for supporting ChIP-seq evidence from ReMap~\cite{coetzee_motifbreakr_2024}. The highest MRS quantile was 3.3-fold enriched for ChIP-seq validation of the TFBS prediction relative to lowest quantile ($p$-value $< 1 \times 10^{-3}$, \hyperref[fig:fig6]{Figure~6c}), confirming that Kosmos’s prioritization aligned with experimentally supported regulatory elements. Across 7,925 GWAS variants, the highest observed MRS was 6.0 for rs9379084. Notably, 6 of the 10 top-scoring SNP–gene pairs corresponded to the same variant (\hyperref[fig:fig6]{Figure~6d}), prompting Kosmos to focus on the rs9379084 locus.

To test whether the protein-coding genes in the rs9379084 locus (RREB1, SSR1, RIOK1, DSP, and SNRNP48) are significantly regulated by the 12 TFs predicted to be disrupted by the same SNP, Kosmos called the \texttt{gseapy} library's \texttt{Enrichr} function for an over-representation analysis. The 5-gene set was tested against several Enrichr TF-target databases (including \texttt{TRRUST\_Transcription\_Factors\_2019}, \texttt{ENCODE\_and\_ChEA\_Consensus\_TFs\_from\_ChIP-X}, \texttt{ChEA\_2022}, \texttt{ENCODE\_TF\_ChIP-seq\_2015}, and \texttt{TF\_Perturbations\_Followed\_by\_Expression}). Enrichr output was then filtered for terms (regulator TFs) matching one of the 12 predicted rs9379084-disrupted TF binding sites. The analysis identified one significant enrichment ($p < 0.05$): the gene set was enriched for targets of ATF3 ($p=0.042$), which regulates two of the five genes (SSR1 and SNRNP48). This significant finding originated from the ATF3 GM12878 hg19 database term. A separate, non-significant ($p=0.31$) database match was also found between ATF3 and RREB (\hyperref[fig:fig6]{Figure~6e}). Finally the synthesis of this finding was: ``In aggregate, rs9379084-A connects a missense-tagged regulatory sequence to increased ATF3 binding, islet-selective chromatin closure, and directionally coherent (though variably powered) expression changes for SSR1 [...], providing a plausible protective pathway that reprograms stress-responsive regulatory circuits in pancreatic islet cells to reduce type 2 diabetes risk'' (\hyperref[fig:fig6]{Figure~6f}). 

Independent validation of this mechanistic hypothesis identified supporting evidence in the pancreas-specific enhancer-target data~\cite{gschwind_encyclopedia_2023} which shows that the missense SNP rs9379084-A overlaps with a cis-regulatory element that is predicted to regulate SSR1 and RREB1 (\hyperref[fig:fig6]{Figure~6g}). Additionally, the transcriptome-wide association analyses (TWAS) hub data~\cite{gusev_integrative_2016} shows that SSR1 is the only gene in the locus with a significant association with T2D (TWAS $|Z| > 5$, \hyperref[fig:fig6]{Figure~6h}), suggesting that genetically-inferred levels of SSR1 are associated with higher risk of T2D. Lastly, previous studies have demonstrated that SSR1 levels are causally linked to circulating insulin in cell~\cite{huang_deficient_2021} and animal models~\cite{li_trap_2025}. 

This discovery demonstrated Kosmos' ability to autonomously develop a target prioritization strategy for likely causal mechanisms in T2D, followed by proposing a specific mechanistic hypothesis that is supported by independent data sources and literature.

\subsection{Kosmos independently develops new methods}

\subsubsection{Discovery 6: Temporal ordering of disease-related events in Alzheimer's Disease}

\begin{figure}[t!p]
\includegraphics[width=\columnwidth]{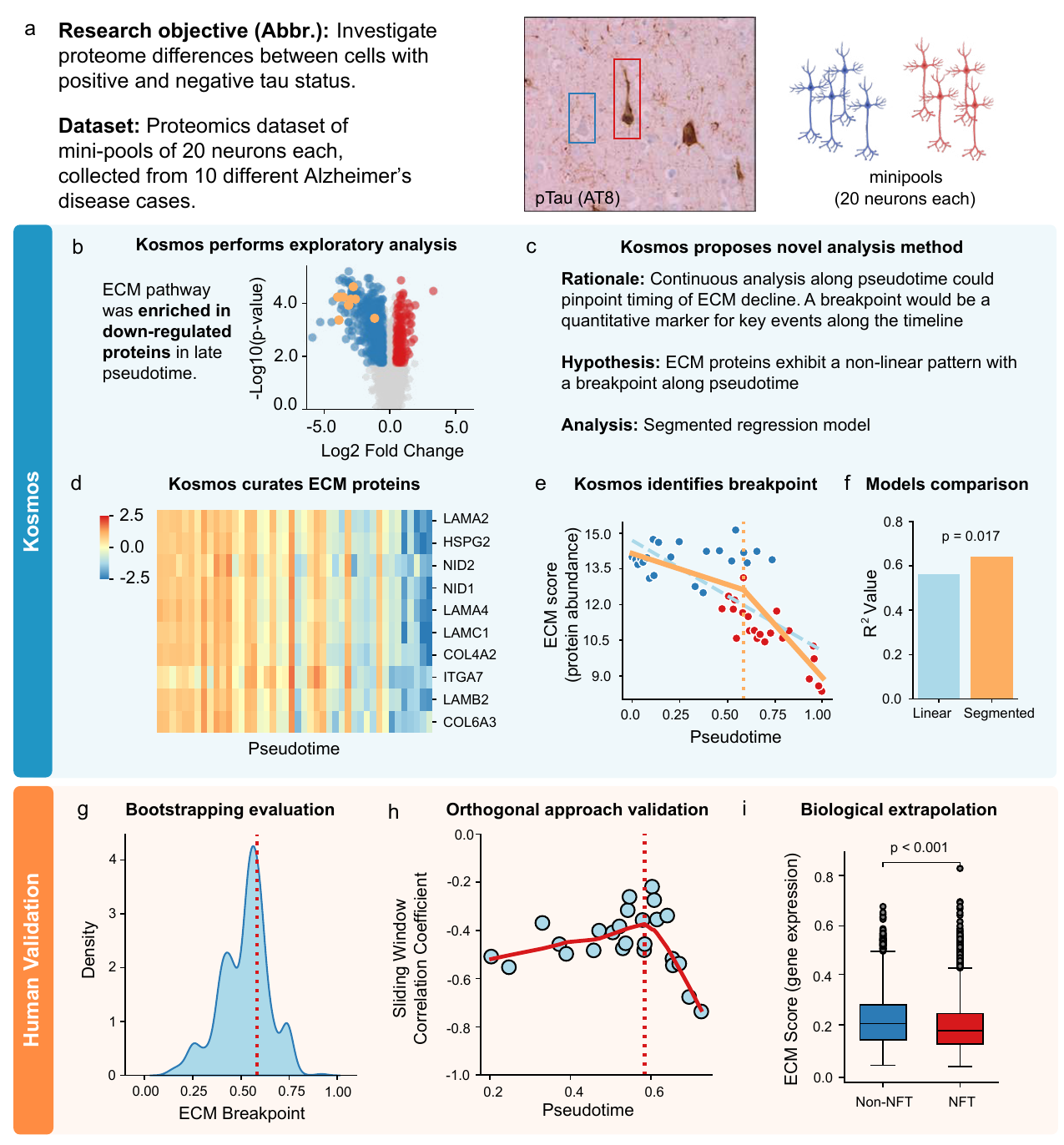}
\caption{\textbf{Kosmos develops temporal ordering method for disease-related events in Alzheimer's Disease.} \textbf{a)} Research objective and dataset description given to Kosmos. Representative image of phosphorylated-tau (AT8) staining showing a positive cell (red) and a neighbouring negative cell (blue). Minipools of 20 positive and negative cells were collected for proteomic analysis. (Diagram created in \url{https://BioRender.com}) \textbf{b)} Volcano plot showing an exploratory analysis by Kosmos comparing early and late pseudotime samples, identifying extracellular matrix (ECM) pathway (yellow dots) as significantly enriched in down-regulated proteins (blue dots). Red dots indicate up-regulated proteins \href{https://platform.edisonscientific.com/trajectories/5d8f512a-5f08-47c1-9326-a0b081e22238}{[Trajectory r17]}. \textbf{c)} Kosmos reasoned on an analytical method to identify timing of pathway decline. \textbf{d)} Heatmap showing levels (z-scored) of ECM proteins curated by Kosmos along pseudotime. \textbf{e)} Scatter plot of the ECM composite score versus pseudotime (red: Tau$+$, blue: Tau$-$) fitted with a segmented regression model (yellow line) and a linear model (blue line). The model identifies a breakpoint (yellow dotted line) at 0.58 units. \textbf{f)} $R^2$ values of linear and segmented models (Davies test - $p$ = 0.017) \href{https://platform.edisonscientific.com/trajectories/03781271-2285-4bf0-9b7e-f481a7360117}{[Trajectory r26]}. \textbf{g)} Stability of breakpoint was assessed by bootstrapping. Shaded area: kernel density estimate. Red dotted line: breakpoint. \textbf{h)} Sliding window correlation between ECM composite score and pseudotime with LOWESS fit reveals an inflection point at ECM breakpoint. \textbf{i)} Validation: ECM score decline is also observed at the transcriptome level in an independent dataset: tau$-$ neurons (non-NFT AD) and tau$+$ (NFT AD) (adj. $p$ = 2.75$\times$10$^{-92}$, pairwise Mann--Whitney U test, Bonferroni-corrected).}
\label{fig:fig7}
\end{figure}

Beyond replicating published findings or executing suggested analyses, Kosmos can propose novel analytical approaches. Here, we demonstrate how Kosmos proposed a data-science-driven approach to pinpoint when a given cellular process was affected in a disease continuum. 

Alzheimer's disease (AD) is the most common form of dementia. Brain pathology reflects the accumulation of extracellular protein aggregates in the form of amyloid plaques and intracellular aggregates of the microtubule-binding protein tau into structures known as neurofibrillary tangles (NFTs). Despite the broad and constitutive expression of tau across the brain, NFTs form in a remarkably selective and stereotyped pattern, affecting distinct neuronal populations and brain regions in a predictable sequence. This phenomenon, known as selective neuronal vulnerability, remains one of the most enigmatic features of diseases with NFT pathology. To investigate the underlying molecular mechanisms of neuronal vulnerability, Foiani \textit{et al.} combined laser-capture microdissection with ultrasensitive mass spectrometry to profile protein abundance in affected neurons (i.e. neurons positive for phosphorylated tau accumulation detected by antibody AT8) alongside neighboring unaffected neurons in the same brain. Prior to providing this data to Kosmos, quality control was performed and samples were ordered on a pseudotime axis closely reflecting intracellular tau accumulation (Foiani \textit{et al.} in preparation).

The primary research objective given to Kosmos was to propose mechanisms contributing to tau accumulation and a temporal sequence of these events (Methods and \hyperref[fig:fig7]{Figure~7a}). Kosmos initially performed a differential abundance analysis by comparing minipools in the earliest 33\% of the pseudotime to those in the latest 33\% (\hyperref[fig:fig7]{Figure~7b}). The down-regulated proteins were enriched in proteins involved in the extracellular matrix (ECM), closely replicating independent human analyses (Foiani \textit{et al.} in preparation) and in agreement with previous studies reporting reduced expression of ECM proteins and upregulation of metalloproteinases~\cite{schmidt_tau_2022}.

As the research objective required Kosmos to propose a sequence of events, the timing of the events would have to be determined. Kosmos rationalized that although the binned analysis showed differences in early and late stages, a continuous analysis could pinpoint the timing of pathway changes more precisely. Kosmos hypothesised that the decline in ECM proteins would exhibit a non-linear pattern with a breakpoint that indicates the timing of ECM failure (\hyperref[fig:fig7]{Figure~7c}). To test this hypothesis, Kosmos first selected 10 ECM proteins and calculated a composite ECM abundance score using the mean abundance of those 10 proteins (\hyperref[fig:fig7]{Figure~7d}). Then, segmented regression models were fitted to the ECM score against pseudotime. The best fit was determined by optimizing for residual sum of squares (RSS) (\hyperref[fig:fig7]{Figure~7e}). The resulting model provided a significantly better fit than a simple linear model (Davies test, $p = 0.017$, \hyperref[fig:fig7]{Figure~7f}), and pinpointed a breakpoint at 0.58 units of pseudotime. Of interest, the proposed breakpoint is later than the transition between tau-negative and tau-positive (around 0.5), suggesting that ECM failure is downstream of tau accumulation. This approach can be expanded towards other affected cellular pathways to reconstruct the sequence of events leading to tau pathology formation.

To assess the computational replicability, we repeated the analysis across multiple independent trials with the data analyst agent, obtaining consistent support for Kosmos' proposed breakpoint (see Methods). Of interest, during this process, the agent proposed alternative computational approaches that further strengthened the finding and were manually validated. The agent implemented a bootstrapping approach (\hyperref[fig:fig7]{Figure~7g}) and an orthogonal breakpoint analysis (by identifying the inflection point of a sliding window correlation, \hyperref[fig:fig7]{Figure~7h}), which both gave the identical conclusions. 

To validate the biological relevance of this finding, we calculated the ECM composite score in the same manner in previously published single-soma transcriptomic data~\cite{otero-garcia_molecular_2022}. A similar decrease of ECM gene expression was observed in this dataset between neurons from AD patients without neurofibrillary tangles (NFT) and those with NFT (\hyperref[fig:fig7]{Figure~7i}), thus suggesting a possible adaptation to tau accumulation. The complete study will be reported in an upcoming manuscript (Foiani \textit{et al.}, in preparation).

This discovery highlighted Kosmos' ability to not only execute standard analyses, such as differential expression or pathway enrichment analysis, but also to reason and propose unconventional methods to address research objectives. While optimizing a segmented model to fit data is prevalent in data sciences~\cite{wagner_segmented_2002}, this approach, to the best of our knowledge, has not been previously applied within a biological context to investigate molecular events along a cellular timeline. 

\subsection{Kosmos makes novel clinical discoveries not identified by human researchers}

\subsubsection{Discovery 7: Mechanism of entorhinal cortex vulnerability in aging}

\begin{figure}[t!p]
\includegraphics[width=\columnwidth]{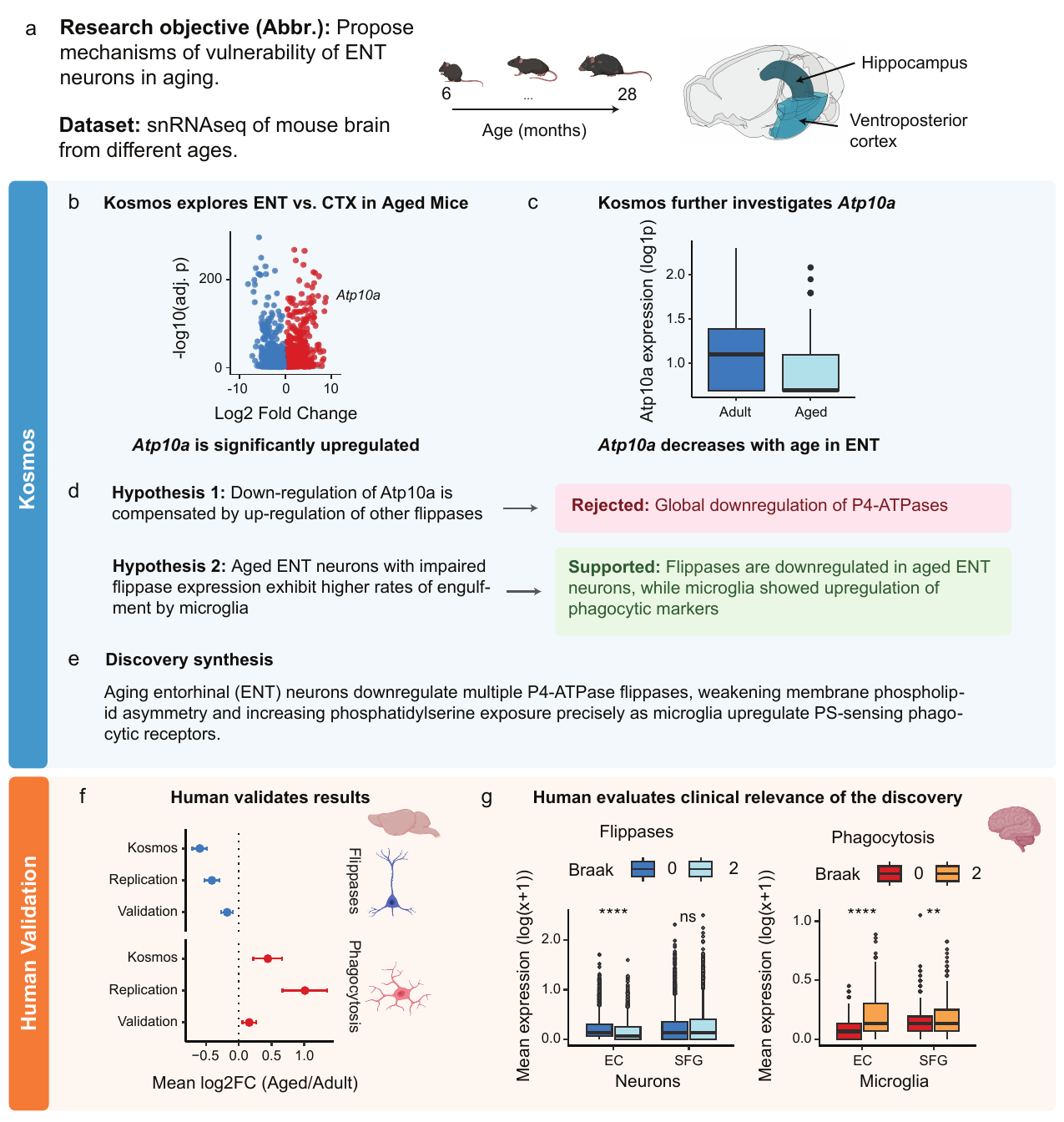}
\caption{\textbf{Kosmos makes novel discovery of entorhinal cortex vulnerability mechanism in aging.} \textbf{a)} Research objective and dataset description given to Kosmos. Mice aged 6 to 28 months were used in this study. 3D brain schematic highlights brain regions (hippocampal formation and ventroposterior cortex, including entorhinal cortex) dissected for nuclei isolation \textbf{b)} Volcano plot showing  differential expression analysis between ENT vs CTX in aged mice, and identifying \textit{Atp10a} as significantly up-regulated. Red: up-regulated genes; blue: down-regulated genes. \href{https://platform.edisonscientific.com/trajectories/5fe55261-ef26-4ad3-838e-02d2c4ce539b}{[Trajectory r2]}. \textbf{c)} Box plot showing that \textit{Atp10a} decreases with age in ENT neurons. Literature search suggested functional and clinical relevance of \textit{Atp10a} to the research objective \href{https://platform.edisonscientific.com/trajectories/be179754-a550-442a-b6d1-72efb4474993}{[Trajectory r32]}. \textbf{d)} Iterative hypothesis exploration. The flowchart shows representative hypotheses generated and tested to arrive at the discovery. \textbf{e)} An excerpt from the discovery report generated by Kosmos relating to the reported discovery. \textbf{f)} Downregulation of flippase and upregulation of microglial activation genes in original Kosmos run, a replication analysis with the same data (replication) and a validation analysis using an independent published dataset (validation). \textbf{g)} Validation of clinical significance in Alzheimer's disease Braak 0 vs 2 stage cases. (left) Flippases expression (\textit{ATP8A1}, \textit{ATP8A2}, \textit{ATP8B1--4}, \textit{ATP9A--B}, \textit{ATP10A--D}, \textit{ATP11A--C}).  (right) Phagocytosis related genes (\textit{CD68}, \textit{TREM2}, \textit{TYROBP}, \textit{C1QA--C}, \textit{CSF1R}, \textit{CX3CR1}, \textit{P2RY12}, \textit{TMEM119}, \textit{HEXB}, \textit{CTSD}, \textit{LAMP1--2}). (Diagrams in a,f,g were created with \url{https://BioRender.com}) }
\label{fig:fig8}
\end{figure}

Here, we demonstrate how Kosmos was able to uncover a clinically relevant mechanism of neuronal aging that was not identified by the researchers who originally created and analyzed the dataset.

Aging is the most common risk factor for dementia but we still do not understand the molecular mechanisms involved. Tau pathology first emerges in specific vulnerable cell-types and brain regions, such as supragranular excitatory neurons from the entorhinal cortex~\cite{mrdjen_basis_2019}. This pattern of early vulnerability is consistently observed in both human patients and corresponding mouse models~\cite{watamura_vivo_2025, fu_selective_2018}, providing a powerful system for investigation. Bourdenx \textit{et al.} (in preparation) hypothesized that investigating the specific mechanisms that make entorhinal neurons vulnerable during aging will provide critical insights into the initial stages of Alzheimer's disease (AD).

Kosmos was directed to identify mechanisms of vulnerability in aging by comparing vulnerable Layer 2/3 entorhinal cortex (ENT) neurons against resilient Layer 2/3 isocortex cortex (CTX) neurons, given single-nuclei RNA-sequencing data derived from brains of mice aged 6--28 months (\hyperref[fig:fig8]{Figure~8a}). Kosmos first conducted a high-level exploration of the transcriptomic dataset. Differential expression analysis in aged mice revealed that \textit{Atp10a} expression was significantly higher in ENT compared to CTX neurons (\hyperref[fig:fig8]{Figure~8b}). Kosmos then conducted a literature review, identifying \textit{Atp10a} as a phospholipid flippase implicated in the survival of glutamatergic neurons and genetically linked to neurodegenerative diseases (OpenTargets). 

Further analysis by Kosmos showed that \textit{Atp10a} expression in ENT was downregulated with age (\hyperref[fig:fig8]{Figure~8c}). Kosmos explored a loss-of-function mechanism, investigating whether the age-related downregulation of \textit{Atp10a} in ENT was compensated by other P4-ATPase flippases. This hypothesis was also refuted, as the analysis revealed a concurrent downregulation of nine other flippase family members (\hyperref[fig:fig8]{Figure~8d}). This widespread loss of flippase activity led Kosmos to propose a novel mechanism of vulnerability: the reduction in flippase activity resulting in increased exposure of phosphatidylserine (PS) on the outer neuronal membrane, which is an established ``eat-me'' signal that triggers microglial phagocytosis~\cite{scotthewitt_local_2020}. Consistent with this model, Kosmos performed subsequent analysis that confirmed a concurrent significant downregulation of flippase genes in neurons and the upregulation of PS-sensing phagocytic pathway genes in microglia, providing a plausible mechanism for the targeted removal of vulnerable neurons in the aging entorhinal cortex (\hyperref[fig:fig8]{Figure~8e}). 

These findings were validated through two independent methods. First, we tested the validity of the finding in an independent dataset generated by another research group~\cite{jin_brain-wide_2025}. Then, we assessed computational replicability by prompting our data analyst agent to confirm the claim made by Kosmos (see Methods) over multiple runs. Both approaches confirmed the observation of an age-dependent collapse of the flippase machinery in the entorhinal cortex jointly with an upregulation of the pro-phagocytosis axis in microglia (\hyperref[fig:fig8]{Figure~8f}). Of note, the validation dataset was single-cell RNA sequencing (as opposed to single-nuclei for the discovery dataset), likely contributing to the observed difference in effect size, with the validation dataset showing more modest changes.

To evaluate the clinical relevance of this discovery, we analyzed an external single-cell RNA sequencing dataset from human AD cases~\cite{leng_molecular_2021}, where a comparable trend was observed (\hyperref[fig:fig8]{Figure~8fg}). We observed a decrease in flippase expression in supragranular neurons from the entorhinal cortex at Braak stage II compared to a cortical-free stage (Braak 0), indicating that such decrease coincides with the appearance of tau pathology in that region. Notably, this decrease was not observed in a later affected region (i.e. superior frontal gyrus, SFG). In addition, microglia from the EC and to a lesser extent the SFG demonstrated concurrent upregulation of phagocytosis markers, confirming the coordinated modulation of this pro-phagocytosis axis. 

While the exposure of PS as an ``eat-me'' signal is a recognized mechanism for cellular and synaptic clearance in the brain~\cite{scotthewitt_local_2020, ruedacarrasco_microgliasynapse_2023}, this discovery establishes a novel mechanism for ENT neuron vulnerability in aging and AD. Flippases are essential for the maintenance of membrane asymmetry, signaling, or vesicular trafficking~\cite{sakuragi_regulation_2023}. Of interest, A$\beta$ oligomers have been shown to induce PS exposure and hyperactivity~\cite{ruedacarrasco_microgliasynapse_2023}. The observed age- and tau-related downregulation of flippases in ENT neurons could further amplify this phenomenon and contribute to the early loss of entorhinal to hippocampal connectivity in AD observed in preclinical patients~\cite{stoub_hippocampal_2006} and indicate novel potential areas for intervention. This discovery demonstrates that, through iterative exploration, Kosmos can identify disease-relevant overlooked mechanisms in existing datasets.

\section{Discussion}

To our knowledge, Kosmos is the first example of an AI Scientist that can carry out months of work in a single run, combining closed-loop literature search, data analysis, and world model updates to autonomously make discoveries in multiple fields. Kosmos does this by using structured world models to manage context between agents, allowing it to deploy hundreds of agent rollouts, write tens of thousands of lines of code, and read thousands of papers to complete a single research objective. Kosmos performs large-scale, unbiased exploration of high-dimensional datasets to successfully reproduce known work (Figures \hyperref[fig:fig2]{2}-\hyperref[fig:fig4]{4}), refine and synthesize existing knowledge (Figures \hyperref[fig:fig5]{5}-\hyperref[fig:fig7]{7}), and make novel discoveries (\hyperref[fig:fig8]{Figure 8}). Because Kosmos uses a dynamically updated world model to deploy two general-purpose scientific agents, Kosmos is able to operate in any domain. While we have demonstrated its utility in metabolomics (\hyperref[fig:fig2]{Figure 2}), materials science (\hyperref[fig:fig3]{Figure 3}), connectomics (\hyperref[fig:fig4]{Figure 4}), statistical genetics (Figures \hyperref[fig:fig5]{5}, \hyperref[fig:fig6]{6}), proteomics (\hyperref[fig:fig7]{Figure 7}), and transcriptomics (\hyperref[fig:fig8]{Figure 8}), we anticipate that Kosmos can be applied to diverse data-rich fields. Finally, each statement in a Kosmos report is supported by a piece of code or primary literature citation. This allows the entire discovery process to be transparent and facilitates independent verification or replication of any finding, as presented for the discoveries here.

\subsection{Scientist-in-the-loop}

Kosmos is designed not to replace human scientists, but to augment and accelerate their work. A Kosmos-integrated pipeline begins with human-generated and -curated high-quality dataset and ends with human interpretation and critical evaluation of the results. Thus, the quality and format of input datasets have a significant impact on the output discoveries. For instance, preliminary runs for discoveries in Figures 2, 5 and 6 yielded different focuses and results depending on the type of input data pre-processing (data not shown). Throughout our testing, collaborators found Kosmos provided the best scientific insight when given clearly labeled, well-formatted, and properly normalized data.

Following a Kosmos run, the scientist’s role shifts to evaluation and interpretation of results. This human oversight is crucial, as independent evaluators noted that Kosmos tends to make excessively strong claims and can sometimes veer in unexpected trajectories (\hyperref[fig:fig1]{Figure 1c}). Together, scientists and Kosmos form a rapid feedback loop where initial human findings feed AI-generated hypotheses which are then refined to guide subsequent searches, ensuring Kosmos' analytical power remains directed toward accurate and meaningful scientific goals.

\subsection{Limitations and Future Work}

Kosmos has several limitations that highlight opportunities for future development. First, although 85\% of statements derived from data analyses were accurate, our evaluations do not capture if the analyses Kosmos chose to execute were the ones most likely to yield novel or interesting scientific insights. Kosmos has a tendency to invent unorthodox quantitative metrics in its analyses that, while often statistically sound, can be conceptually obscure and difficult to interpret. Similarly, Kosmos was found to be only 57\% accurate in statements that required interpretation of results, likely due to its propensity to conflate statistically significant results with scientifically valuable ones. Given these limitations, the central value proposition is therefore not that Kosmos is always correct, but that its extensive, unbiased exploration can reliably uncover true and interesting phenomena. We anticipate that training Kosmos may better align these elements of “scientific taste” with those of expert scientists and subsequently increase the number of valuable insights Kosmos generates in each run. 

Second, identifying valuable discoveries Kosmos made is a time-intensive process that relies on human scientists with significant domain expertise. An average Kosmos report contains 3-4 discovery narratives, and each discovery narrative contains ~25 claims based on ~8-9 agent trajectories. While the provenance of each claim is traceable, there exists no automated method to reliably evaluate if a claim is accurate, novel, and significant. Thus, although Kosmos performs data-driven discovery at scale, identifying meaningful discoveries at scale remains a challenge. 

Finally, the current implementation of Kosmos faces several technical limitations. Kosmos can only manage datasets up to approximately 5GB, and does not currently excel in the analysis of raw data, such as images or raw sequencing files. Kosmos cannot autonomously access publicly available data from external sources to use as a reference or in orthogonal validations.   Furthermore, as Kosmos is stochastic, multiple independent runs may not consistently converge on the same discoveries. The system's research directions are currently sensitive to the phrasing of the research objectives. Finally, the current implementation does not allow for scientists to interact with Kosmos in intermediate cycles, limiting opportunities for scientists to nudge Kosmos down fruitful research avenues. We anticipate addressing these challenges in future work. 

\subsection{Conclusion}

Taken together, our results demonstrate an AI scientist that can autonomously conduct extended research investigations that produce discoveries validated by domain experts across multiple scientific fields. The world model enables coordination of parallel agent trajectories at scales that would require months of human effort, while maintaining complete traceability of all scientific claims. With further training, Kosmos has the potential to significantly scale data-driven discovery.

\section{Methods}
\label{sec:methods}

\subsection{Figure 1 Methods}

\subsubsection{Expert evaluations of Kosmos accuracy}

Expert scientists were asked to evaluate a sample of Kosmos statements to determine if it is accurate and/or reproducible. Initially, these experts were asked to mark the statements as "SUPPORTED", "REFUTED", or "UNSURE". The instructions for evaluation are shown in \hyperref[sec:supp_eval_overview]{Supplementary Information 3}. Upon submission, all answers that were marked as "UNSURE" were reviewed to determine if the expert was unable to perform the analysis or if the statement was malformed or insufficient. In the latter case, the statement was clarified or additional relevant information was provided, and the evaluator was asked to determine definitively if the statement was "SUPPORTED" or "REFUTED". Examples of both "SUPPORTED" and "REFUTED" evaluations for Data Analysis, Literature Review, and Interpretation statements are provided in \hyperref[sec:supp_eval_answers]{Supplementary Information 4}.

\subsection{Figure 2 Methods}

\subsubsection{\textit{in vivo} experiment}

All experiments used adult KOR-Cre mice (9–10 weeks old) maintained on a 12-hour light/dark cycle with \textit{ad libitum} access to food and water. All procedures were approved by the Washington University Animal Studies Committee, and experimenters were blinded to the treatment group.

For chemogenetic activation, mice received bilateral stereotactic injections of AAV8-hSyn-DIO-hM3Dq-mCherry (or control AAV8-hSyn-DIO-mCherry) into the medial preoptic area (POA). Viral expression was verified at least three weeks later by a test dose of clozapine-N-oxide (CNO; 0.5~mg/kg, intraperitoneal). Animals failing to show a decrease in core body temperature of at least 3\textdegree C were excluded.

Seventeen validated mice were then assigned to one of three experimental conditions. Control mice received the control virus and were housed at room temperature ($\sim$22\textdegree C). Hypothermic-POA$\textsuperscript{KOR+}$ mice received the DREADD virus and were housed at room temperature ($\sim$22\textdegree C) to allow the development of a torpor-like hypothermic state. Normothermic-POA$\textsuperscript{KOR+}$ mice also received the DREADD virus but were housed at 36\textdegree C to prevent cooling and maintain normothermia.

Thirty minutes after CNO administration, animals were euthanized by live decapitation. Whole brains were rapidly removed, flash-frozen in liquid nitrogen, and stored at $-80$\textdegree C. Polar and lipid metabolites were extracted as described previously~\cite{kamal_preoptic_2025}: tissue was homogenized in a 2:2:1 acetonitrile:methanol:water mixture (40~$\mu$ L per mg tissue) using a bead mill, incubated at $-$20\textdegree C for 1 hour, and centrifuged at 10,000~rpm for 10 minutes at 4\textdegree C. The supernatant was transferred to autosampler vials and stored at $-$80\textdegree C until analysis.

\subsubsection{Metabolomic profiling and data preprocessing}

Metabolomic profiling was performed on a Thermo Scientific Vanquish Flex UHPLC system coupled to a Thermo Scientific Orbitrap ID-X mass spectrometer. Polar metabolites were separated on a HILICON iHILIC-(P) Classic HILIC column, and lipids were separated on an Acquity UPLC HSS T3 column using the solvent gradients described in the original study. Data were acquired in both positive and negative ion modes.

Raw spectra were processed in XCMS, Compound Discoverer (v3.3), and Skyline for peak detection, retention-time alignment, and compound identification. The resulting normalized peak-intensity matrix served as the input dataset for Kosmos analyses. Statistical comparisons between hypothermia and control groups were performed using unpaired \textit{t}-tests (Figure 2B).

The results of Kosmos run were validated by comparing to the preprinted statistical results.

\subsection{Figure 3 Methods}

To investigate how the ambient environment during the synthesis of FAPbI$_3$ perovskite thin films influences solar cell performance, two custom-built environmental enclosures were used to enable independent control of ambient conditions during the spin-coating and thermal annealing steps. In the spin-coating chamber, we independently regulated temperature, absolute humidity (AH), and the dimethylformamide (DMF) solvent partial pressure (SPP) via CDA, a humidifier/water boiler, and a DMF bubbler, with resistive heating and local sensors for feedback. In the annealing chamber, AH was actively controlled; temperature was set by the hotplate (no separate chamber temperature control), although it was measured and included in our regressor. Perovskite films for n--i--p solar cell devices were fabricated using a one-step antisolvent spin-coating method followed by thermal annealing. The resulting devices were characterized for short-circuit current density (JSC), open-circuit voltage (VOC), fill factor (FF), and power conversion efficiency (PCE). To efficiently explore the multidimensional process space, we employed Bayesian Optimization (BO) in iterative learning cycles. Experimental results from each cycle informed the algorithm’s suggestion for the next set of synthesis conditions. Full experimental details can be found in Liu et al., 2025~\cite{liu_disentangling_2025}.

The results of Kosmos run were validated by comparing to the preprinted figures and reported trends.

\subsection{Figure 4 Methods}

All data analyzed in this study originate from a preprint~\cite{piazza_physical_2025} that integrated eight publicly available connectome reconstructions, providing single-neuron--resolution synaptic connectivity across five species (\textit{C. elegans}, \textit{Drosophila} larva, adult \textit{Drosophila}, zebrafish, mouse, and human). The datasets encompass the complete \textit{C. elegans} connectome~\cite{white_structure_1986, winding_connectome_2023}, and adult \textit{Drosophila} central nervous system~\cite{dorkenwald_neuronal_2024, scheffer_connectome_2020} and ventral nerve cord~\cite{takemura_connectome_2024} reconstructions, the larval zebrafish brainstem~\cite{vishwanathan_predicting_2024}, a millimeter-scale volume of mouse visual cortex~\cite{schneider-mizell_inhibitory_2025}, and a human temporal cortex sample~\cite{shapson-coe_petavoxel_2024}. The preprint harmonized these datasets by standardizing neuron and synapse formats and rescaling spatial coordinates to account for imaging resolution and section thickness, allowing cross-species comparison of neuron morphology and connectivity. From this preprocessed data, for which full details are provided in the preprint~\cite{piazza_physical_2025}, we downloaded morphological details of fully reconstructed neurons and all synapses associated with them. Specifically, we utilized neuronal size $L$ (total path length of each neuron’s reconstructed skeleton), synapse count $S$ (total number of pre- and post-synaptic contacts per neuron), and local synapse density $\rho$ (estimated using a random-walk approach that samples local synapse-per-micron density around each synapse), and neuronal degree $k$ (number of unique partner neurons per cell).

The results of Kosmos runs were validated by comparing to the preprinted figures and reported fit parameters.

\subsection{Figure 5 Methods}

\subsubsection{Data description and preprocessing}
Publicly available datasets were utilized. Genetic variants that influence protein levels (known as cis-protein quantitative trait loci or cis-pQTLs) were used as the exposure data. GWAS summary statistics from plasma proteomics~\cite{coetzee_motifbreakr_2024}. These genetic variants were located within 500 kilobases of the protein-coding gene, a region defined as "cis" in this context. To ensure the genetic instruments were independent from each other, linkage disequilibrium (LD) clumping was performed to retain only variants with minimal correlation ($r^2 < 0.01$). Independent variants were retained after LD-based clumping ($r^2 < 0.01$). Myocardial T1 relaxation time GWAS summary statistics from Nauffal \textit{et al.} was used as the outcome data~\cite{nauffal_genetics_2023}.

\subsubsection{Discovery validation}

To validate Kosmos results, Mendelian Randomization with the inverse-variance weighted (IVW) method was conducted independently on the same dataset. The MR workflow was implemented using the MendelianRandomization R package~\cite{yavorska_mendelianrandomization_2017}. The number of SNPs used as instruments was recorded for each protein-outcome pair. Results were considered statistically significant if they achieved a p-value below the Bonferroni-corrected threshold (0.05 divided by the number of tests performed within each analysis type).

\subsection{Figure 6 Methods}

\subsubsection{Data description and preprocessing}

Publicly available datasets were utilized. The 10x multiome data from the Wang et al study~\cite{wang_integrating_2023} was accessed via GEO (accession IDs:  GSE124742 and GSE164875). The T2D GWAS summary statistics were accessed on the DIAGRAM consortium website (https://diagram-consortium.org/downloads.html). The snRNAseq and snATAC-seq data from all of the available cell clusters were processed as previously described~\cite{wang_integrating_2023}. Cell clusters with less than 1000 cells were excluded from the analysis. The R package motifbreakR~\cite{coetzee_motifbreakr_2024} was used to predict transcription factor binding sites (TFBS) that overlap with the fine-mapped GWAS variants. The ReMap lookup function was used when running motifbreakR to include validated predicted TFBS with ChIP-seq validation. The ATAC-seq and the snRNAseq counts were converted into pseudobulk counts using harmony~\cite{wang_integrating_2023}. The ATAC peak ~ transcript level  associations were conducted using a linear regression model which adjusted for sex, BMI, and disease status (since we were interested in a disease-agnostic association between ATAC peaks and transcript levels). eQTL and pQTL summary stats were provided from pancreas-specific GTEx data~\cite{the_gtex_consortium_gtex_2020} and blood-specific proteome data~\cite{sun_plasma_2023}, respectively. Lastly, we queried Gene Ontology for annotations for all of the putative target genes and the transcription factors that were predicted to have their binding sites disrupted by the GWAS variants. The resulting table contained 68K SNP-TF pairs from 7,925 unique GWAS SNPs. A data dictionary that explained the significance of each column was also generated. The table and the data dictionary were both used as input for Kosmos.  

\subsubsection{Discovery validation}

The IGV plot with the enhancer-to-gene (E2G)  tracks were downloaded on 10/24/2025 here: \url{https://e2g.stanford.edu/variant/6_7231610_G_A}
 
The TWAS reflection plot was generated using Finch: \url{https://platform.edisonscientific.com/trajectories/aa8f203a-a27c-454e-b4be-25c1ed67c965}

\subsection{Figure 7 Methods}

\subsubsection{Data collection and preprocessing}

Formalin-fixed paraffin embedded (FFPE) sections (8~$\mu$m) were collected from 10 cases on slides compatible with laser-capture microdissection. Sections were stained for phosphorylated Tau (Ser202/Thr205 - AT8) and minipools of 20 positive and neighboring negative cells were collected and submitted to ultrasensitive mass spectrometry. Following data collection, standard quality control and imputation were used. Minipools were ordered on a pseudotime axis using the palantir package~\cite{setty_characterization_2019}. The complete method and dataset will be reported in an upcoming publication (Foiani, et al., \textit{in preparation}).

\subsubsection{Initial discovery}

The dataset and prompt (Supplementary Table 1) were provided to Kosmos, which was run for 35 iterations. Discovery reports were generated after iterations 8 and 35. Upon review, the discoveries from iteration 8 presented a more cohesive narrative. To refine the final output, we manually-curated these discoveries by writing new titles and summaries based on the iteration 8 report. A final, consolidated discovery report was then generated using these curated descriptions and the complete set of trajectories from all 35 iterations.

\subsubsection{ECM composite score}

Pathway enrichment analysis of down-regulated proteins identified a significant overrepresentation of proteins involved in pathways such as extracellular matrix (ECM). Protein-Protein-Interaction network analysis identified fibronectin (FN1) as a high centrality protein linking ECM to other affected pathways. The top ten ECM proteins with highest correlation with FN1 (LAMA2, HSPG2, NID2, NID1, LAMA4, LAMC1, COL4A2, ITGA7, LAMB2, COL6A3) were considered the most tightly co-regulated ECM protein networks. Their mean expressions were defined as “composite ECM score” for downstream analysis.

\subsubsection{Discovery Validations}

For replication runs, the data analysis agent was provided the same dataset and prompted to test Kosmos claim: \textit{“An ECM composite score (mean of LAMA2, HSPG2, NID2, NID1, LAMA4, LAMC1, COL4A2, ITGA7, LAMB2, COL6A3) declined nonlinearly with pseudotime, with a significant breakpoint at 0.5754 where decline accelerated 3.47-fold”}. This process was repeated over 5 independent trajectories, which all strongly supported the claim.

To validate the identified breakpoint, we performed a sliding window correlation analysis. Pearson correlation was calculated in a sliding window of 20 samples, with step size of one sample along the pseudotime axis. A Locally Weighted Scatterplot Smoothing (LOWESS) fit to the resulting correlation coefficients revealed a distinct inflection point at a pseudotime of 0.57, which coincided with the Kosmos-defined breakpoint. This method was adapted from a data analysis trajectory from another discovery of the same Kosmos output.

For further validation, we used a publicly available single-soma transcriptomics dataset from neurofibrillary tangles (NFT) bearing and non-NFT bearing samples from AD and controls37. The data analysis agent was prompted to “Calculate a ECM score using the genes LAMA2, HSPG2, NID2, NID1, LAMA4, LAMC1, COL4A2, ITGA7, LAMB2, COL6A3. Compare the ECM score between control, non-NFT bearing and NFT bearing samples.” All 5 independent trajectories supported the initial discovery where ECM scores decrease from control to non-NFT to NFT bearing samples.

\begin{table}[ht]
\centering
\caption{Links to validation data analysis trajectories}
\begin{tabular}{p{0.35\textwidth} p{0.60\textwidth}}
\hline
\textbf{Analysis Type} & \textbf{Trajectory Links} \\
\hline
Replication (original data) & 
\url{https://platform.edisonscientific.com/trajectories/4d948f9c-7402-48c5-a0f3-41daff09f737} \newline
\url{https://platform.edisonscientific.com/trajectories/d40c1133-2e28-452f-b92e-75d4ba87d2e4} \newline
\url{https://platform.edisonscientific.com/trajectories/13fee7fd-b5ef-4a61-bebf-cb2dcec678f0} \newline
\url{https://platform.edisonscientific.com/trajectories/627619c1-b726-4c4c-bbed-743e55f456b6} \newline
\url{https://platform.edisonscientific.com/trajectories/d8ef5055-113d-4fb4-91a0-42a9e53e7c98} \\
\hline
Validation (Single soma RNA-seq) & 
\url{https://platform.edisonscientific.com/trajectories/e2ebf0d3-74b3-484a-a821-906d23109c94} \newline
\url{https://platform.edisonscientific.com/trajectories/9e50daec-4670-45ce-bb20-9e1e6ff6e5fb} \newline
\url{https://platform.edisonscientific.com/trajectories/06ed2bf0-9ecd-4a54-bec9-40f4826c42bb} \newline
\url{https://platform.edisonscientific.com/trajectories/68980a3b-5510-43bf-8f81-68a5d2367a37} \newline
\url{https://platform.edisonscientific.com/trajectories/51d8ac2b-ae00-4bd6-a595-5696faec19b7} \\
\hline
Sliding window correlation & 
\url{https://platform.edisonscientific.com/trajectories/5b0c9346-0070-45a4-b058-fffb77aef062} \\
\hline
\end{tabular}
\label{tab:trajectories}
\end{table}

\subsection{Figure 8 Methods}

\subsubsection{Data collection and preprocessing}

To characterize aging-associated changes in the mouse brain, the ventral-posterior part of the cortex (i.e. containing entorhinal cortex) and the hippocampal formation were dissected out from and prepared for single-nuclei multiome analysis (n=23 - Age range: 6months - 28months). Both RNA and ATAC libraries were sequenced with a target of 30,000 and 50,000 reads per nucleus, respectively. Reads were demultiplexed to FASTQ files using bcl2fastq and then mapped to the mouse reference genome (mm10). UMIs per genes were counted using cellranger-arc using the –include-intron parameter. Count matrices were processed in R and Python for quality control and filtering. Cell-type annotation was performed using RNAseq data by combining alignment with the Allen Brain Atlas mouse brain taxonomy~\cite{yao_high-resolution_2023} and de-novo clustering, and then extended to the ATACseq data. Note that only RNAseq data was used in Kosmos. The complete dataset will be made available in a separate publication (Bourdenx et al., \textit{in preparation}).

\subsubsection{Discovery Validations}

For replication runs, the data analysis agent was provided the same dataset and prompted to test Kosmos claim: \textit{“The core hypothesis is that an integrated "pro-phagocytic axis" is activated in the aged entorhinal cortex, involving coordinated changes in both neurons and microglia. L2/3 IT ENT neurons exhibit a systemic flippase collapse, implying increased surface "eat-me" signals. Simultaneously, microglia become primed for phagocytosis, coupled with downregulation of the "don't-eat-me" signal.”}. This process was repeated over 5 independent trajectories, which all strongly supported the claim.

For further validation, we used a previously published independent dataset of single cell (as opposed to single nuclei in the discovery dataset) RNA sequencing from mouse whole brain ageing41. The data analysis agent was provided a subsetted dataset only containing ENT, CTX neurons and microglia (downloaded from: \url{https://assets.nemoarchive.org/dat-61kfys3}) with same prompt as before, over 5 independent trajectories, which also strongly supported the initial discovery.

Finally, to assess the clinical relevance of the discovery, we used a publicly available single nuclei RNA sequencing dataset from human AD in an early (EC) and late (SFG) affected regions at various stages (Braak 0, II, and VI). In order to capture the earliest changes in flippase expression, we focused on the comparison between Braak 0 and II, which is the first stage when tau pathology appears in this region. The dataset was subset for supragranular populations (EC:Exc.s1, EC:Exc.s2, EC:Exc.s4, SFG:Exc.s4 and SFG:Exc.s2). The same prompt was used, 4 out of 5 independent trajectories supported the initial discovery.

\begin{table}[ht]
\centering
\caption{Links to validation data analysis trajectories}
\begin{tabular}{p{0.35\textwidth} p{0.60\textwidth}}
\hline
\textbf{Analysis Type} & \textbf{Trajectory Links} \\
\hline
Replication (original data) & 
\url{https://platform.edisonscientific.com/trajectories/6b22ad5a-06f1-4375-903a-0aff5179ce3b} \newline
\url{https://platform.edisonscientific.com/trajectories/152952fb-71a5-4d49-a44d-222142116e0c} \newline
\url{https://platform.edisonscientific.com/trajectories/024b969f-fc41-4c69-83ef-4f493c9b0a9d} \newline
\url{https://platform.edisonscientific.com/trajectories/9b31fb26-beb0-46f3-a93f-f5613ecfa646} \newline
\url{https://platform.edisonscientific.com/trajectories/205e0b9d-d3e7-40b6-9e8a-5bb190763fca} \\
\hline
Validation (mouse whole brain scRNAseq) & 
\url{https://platform.edisonscientific.com/trajectories/7c53ea61-f52a-48b6-839d-0c935ea37d3e} \newline
\url{https://platform.edisonscientific.com/trajectories/9a60b3fe-8ddb-455c-902b-3950c8cbe44c} \newline
\url{https://platform.edisonscientific.com/trajectories/77804eec-6c99-4c76-a165-efa9bf30683f} \newline
\url{https://platform.edisonscientific.com/trajectories/efd47388-34b6-4573-a9b7-f8e14e5175af} \newline
\url{https://platform.edisonscientific.com/trajectories/674572eb-cef9-49c1-9279-fb6c4f7021c9} \\
\hline
Validation (human AD snRNAseq) & 
\url{https://platform.edisonscientific.com/trajectories/72bcc9f2-4a53-468a-93ba-3e22065616ab} \newline
\url{https://platform.edisonscientific.com/trajectories/48c87183-3056-4d63-924b-534983ee3ca8} \newline
\url{https://platform.edisonscientific.com/trajectories/1d007ced-cefb-42ff-8f80-6bfe2da5bb9b} \newline
\url{https://platform.edisonscientific.com/trajectories/08197da6-5d82-4849-9b36-7b44221b7f23} \newline
\url{https://platform.edisonscientific.com/trajectories/11d414a9-a95e-4596-af46-3f6aec0a4f7e} \\
\hline
\end{tabular}
\label{tab:trajectories2}
\end{table}

\section{Data availability statement}

Data for Figures 2-4 are available from respective preprints Piazza \textit{et al.}, 2025~\cite{piazza_physical_2025}, Liu \textit{et al.}, 2025~\cite{liu_disentangling_2025} and Kamal \textit{et al.}, 2025\cite{kamal_preoptic_2025}. Figure 5 utilized publicly available datasets, with plasma proteome GWAS available at \url{https://www.synapse.org/Synapse:syn51365303} and myocardial T1 relaxation GWAS available at \url{https://cvd.hugeamp.org/dinspector.html?dataset=Nauffal2024_Fibrosis_EU&phenotype=MyocardialT1}. For Discovery 5, the 10x multiome data from the Wang \textit{et al.}~\cite{wang_integrating_2023} can be accessed via GEO (Accession IDs: GSE124742 and GSE164875). The GWAS summary statistics are available on the DIAMANTE consortium website. Data for Discovery 6 will be available in a forthcoming publication (Foiani \textit{et al.}, \textit{in preparation}). Similarly, data for Discovery 7 will be made available in a separate publication (Bourdenx \textit{et al.}, \textit{in preparation}).

All Kosmos reports for the discoveries reported here can be found in Supplementary Data 1-7 in pdf format, and on Edison platform (\hyperref[tab:kosmos_reports_links]{Supplementary Table 1}). Jupyter notebooks supporting data analysis and literature citations can be found within the report accessible through the hyperlinks. All scripts used for data analysis and figure generation are available on GitHub at \href{https://github.com/EdisonScientific/kosmos-figures}{https://github.com/EdisonScientific/kosmos-figures}.

\clearpage

\section{Statement of contributions}

L.M. and A.Y. conceived and designed the overall project. L.M. and M.M.H. supervised the project. L.M. led the overall design and engineering of the core system. B.C. led the design of the discovery review system and evaluations of Kosmos system performance. B.C., S.N., A.D.W., L.M., and A.B. contributed to investigating Kosmos' behavior and performance, research, software development and code optimization. T.N., M.S., M.C., A.D.W., and E.M.G. developed and maintained the platform infrastructure for the project. A.D.W. developed the world model concept. L.M., A.Y. and M.M.H. led coordination across collaborating research groups. M.B., E.C.L., D.L.B., N.E., A.S., S.R., B.G., M.F., A.K., and L.P.S. provided, curated, and pre-processed the datasets used for the case studies. M.B., E.C.L., D.L.B., N.E., B.G., and A.S. designed the case study methodologies and evaluated Kosmos-generated discoveries. 
A.Y., M.B., A.S., A.T.W. and S.R. conducted discovery validations. A.Y., M.B., A.S., F.C., B.G., and M.M.H. provided critical user feedback during the system's development cycle. B.G., D.L.B., E.C.L., M.B., K.F.R., S.Z., T.C.O., M.E.O., K.J.Z., and M.M.H. provided scientific evaluation of Kosmos output. B.C. and J.L. coordinated with scientific experts for evaluation. A.Y., B.C., L.M., M.B., A.T.W., E.C.L., D.L.B., N.E., A.S., and M.M.H. prepared the initial draft of the manuscript. A.Y., M.B., B.C., A.S., A.T.W., E.C.L., D.L.B., and N.E. prepared the figures and tables. All authors contributed to the final version of the manuscript. A.D.W. and S.R. supervise research at Edison Scientific, Inc.

\section{Funding information}

Work undertaken by R.J.B., E.C.L., A.K., and L.P.S. is supported by the National Institute of Neurological Disorders and Stroke of the National Institutes of Health under award number R01 NS133365 (to E.C.L.). Work undertaken by K.E.D., M.F. and M.B. is supported by the UK Dementia Research Institute through UK DRI Ltd, principally funded by the Medical Research Council, and by funding from the Cure Alzheimer’s Fund (K.E.D. and M.B.). N.E. and T.B. acknowledge a CIFAR catalyst award through the Accelerated Decarbonization Program and the Toyota Research Institute’s Synthesis Advanced Research Challenge (SARC) challenge. T.R. is supported by the UK EPSRC grant EP/Y037200/1.

\clearpage

\bibliographystyle{unsrt}
\bibliography{references}

\clearpage
\appendix
\section*{Supplementary Information}

\subsection*{Supplementary Table 1: Hyperlinks to Kosmos reports}

\begin{table}[ht]
\centering
\label{tab:kosmos_reports_links}
\begin{tabular}{p{0.45\textwidth} p{0.55\textwidth}}
\toprule
\textbf{Discovery} & \textbf{Hyperlinks} \\
\midrule
1: Nucleotide metabolism in hypothermia & 
\url{https://platform.edisonscientific.com/kosmos/c4bdef64-5e9b-43b9-a365-592dd1ed7587} \\
\addlinespace 
2: Determinant of perovskite solar-cell failure & 
\url{https://platform.edisonscientific.com/kosmos/1fdbf827-be65-4d97-9b66-bf0da600091a} \\
\addlinespace
3: Log-normal connectivity in neural networks & 
\url{https://platform.edisonscientific.com/kosmos/4fb3fbdb-c449-4064-9aa6-ff4ec53131d8} \\
\addlinespace
4: SOD2 as driver of myocardial fibrosis & 
\url{https://platform.edisonscientific.com/kosmos/c6849232-5858-4634-adf5-83780afbe3db} \\
\addlinespace
5: Protective variant of SSR1 in type 2 diabetes & 
\url{https://platform.edisonscientific.com/kosmos/abac07da-a6bb-458f-b0ba-ef08f1be617e} \\
\addlinespace
6: Temporal ordering in Alzheimer’s disease & 
\url{https://platform.edisonscientific.com/kosmos/a770052b-2334-4bbe-b086-5149e0f03d99} \\
\addlinespace
7: Mechanism of neuron vulnerability in aging & 
\url{https://platform.edisonscientific.com/kosmos/28c427d2-be31-48b5-b272-28d5a1e3ea5c} \\
\bottomrule
\end{tabular}
\end{table}

\newpage
\subsection*{Supplementary Information 1: Kosmos inputs for each discovery}
\label{sec:supp_inputs}

\vspace{1em}

\noindent \textbf{Discovery 1} \par
\noindent \textbf{Research Objective} \par
The brain has long been conceptualized as a network of neurons connected by synapses. However, attempts to describe the connectome using established network science models have yielded conflicting outcomes, leaving the architecture of neural networks unresolved. Here, we aggregate Degree Length (\textmu m), Synapses (count), and Density (synapses/\textmu m, a local value, not Synapses/Length, except in the case of C. elegans, where we could not measure it) of neurons from eight experimentally mapped connectomes. Consider how these measures relate from a neuroscientific perspective and neurodevelopment time course in order to propose universal principles among them. Investigate similarities and differences across connectomes and species in the distributions that characterize these metrics, and, if possible, link metrics to each other in a cohesive narrative. \par
\noindent \textbf{Data Description} \par
Multiple *\_AggregatedData.csv files, each representing one connectome (“Animal”):
 Celegans\_AggregatedData.csv (C. Elegans, which does not have local density data), Larva\_LocalDensity\_AggregatedData.csv (Fruit Fly Larva), Hemibrain\_LocalDensity\_AggregatedData.csv (Half of a Fruit Fly’s CNS), MANC\_LocalDensity\_AggregatedData.csv (Fruit Fly Ventral Nerve Cord), FlyWire\_LocalDensity\_AggregatedData.csv (Fruit Fly CNS), Zebrafish\_LocalDensity \_AggregatedData.csv (local reconstruction of zebrafish brain), MM3\_LocalDensity\_AggregatedData.csv (local reconstructions of mouse cortex), H01\_LocalDensity\_AggregatedData.csv (local reconstructions of human cortex).
Schema (exact column names): CellID, Length, Synapses, LocalDensity (absent for c. Elegans), and Degree \par

\vspace{1em}

\noindent \textbf{Discovery 2} \par
\noindent \textbf{Research Objective} \par
The study objective is to identify the impacts of various ambient environmental parameters (solvent partial
pressure, absolute humidity, ambient temperature) on perovskite device efficiency, during a synthesis process
consisting of two sequential steps: a spin coating step, followed by a thermal annealing step. \par
\noindent \textbf{Data Description} \par
\# Description
A structured dataset combining environmental sensor measurements from separate spin coating (absolute humidity, ambient temperature, solvent vapor partial pressure) and thermal annealing (absolute humidity, ambient temperature) enclosures, with perovskite solar-cell device performance metrics open-circuit voltage (VOC), short-circuit current (JSC), fill factor (FF), and power conversion efficiency (PCE). Environmental data were recorded across multiple fabrication rounds using custom-built, feedback-controlled enclosures with calibrated sensors and actuators. Inputs: Environmental sensor data includes two copies of each sample (redundancy).

\noindent \# Data Processing Instructions \par
\noindent \#\# Load Raw Data \par
Read all Excel files in the folder Inputs – environmental sensor data. These are time-dependent readings of the sensor data.
There are multiple tabs in each Excel file, labeled by sample ID number and process step (spin coating or thermal annealing).
There are two synthesis steps for each device: a spin coating step, where perovskite precursors dissolved in dimethylformamide (DMF) are spun coat into a wet thin film, and a thermal annealing step, where the sample is placed on a hotplate and annealed.
Note the environmental sensor data records the temperature of the ambient gas, not the temperature of the hotplate. The ambient gas temperature in the thermal annealing stage was measured, but not independently controlled, because it was dominated by the hotplate.
Read all text files in the folder Outputs - full current-voltage curves. Treat these as raw illuminated J--V (current density vs. voltage) data directly from the solar simulator. \par
\noindent \#\# Load Summary Data \par
Read the file Summary table.xlsx, which nominally contains average values of the sensor data and extracted parameters from the current-voltage curves.
For each current-voltage curve, the following device metrics were extracted: open-circuit voltage (VOC), short-circuit current (JSC), fill factor (FF), power conversion efficiency (PCE). \par
\noindent \#\# Handle Fill Factor Discrepancies \par
Be aware that Kosmos previously miscalculated fill factors (FF) from raw J-V curves.
If FF values differ between the raw curve-derived results and the summary table, default to the summary table values.
Only override this rule if the root cause of the discrepancy can be explicitly identified and justified through reproducible computation. \par
\noindent \#\# Environmental Sensor Precision \par
For all environmental sensor data, determine the minimum step size (i.e., smallest nonzero difference between consecutive values).
Treat this as the measurement precision.
Example: if the smallest increment in temperature readings is 0.1 °C, set temperature sensor resolution = 0.1 °C. \par

\vspace{1em}

\noindent \textbf{Discovery 3} \par
\noindent \textbf{Research Objective} \par
Given that Preoptic area (POA) Kappa Opioid Receptor (KOR+) activation induces a torpor-like, hypothermic and hypometabolic state that is  cerebroprotective only under hypothermic conditions, identify the specific metabolic adaptations that may support this effect. \par
\noindent \textbf{Data Description} \par
Polar (Brain\_Polar\_C\_ST\_NT\_245\_2024.csv) and lipid (Brain\_Lipids\_C\_ST\_NT.csv) untargeted metabolomic screen of brain tissue from mice.  The data is LC/MS peak intensities for three groups: control (N=6), hypothermia (N=6) and normothermia (N=5). The hypothermic group is mice with chemogenetic activation of POA KOR+ neurons. The hypothermic group was allowed to undergo the full drop in body temperature (torpor-like hypothermia).  The normothermia group has the same neuronal activation as above, but animals were housed at elevated ambient temperature to clamp core temperature near normal.  The normothermia group isolates the effect of neural activation without hypothermia, showing whether protection depends on temperature.  Controls received control treatment without POA KOR+ activation and served as baseline comparison for both hypothermia and normothermia groups. \par

\vspace{1em}

\noindent \textbf{Discovery 4} \par
\noindent \textbf{Research Objective} \par
Identify causal proteins in myocardial fibrosis with T1 as a proxy and propose mechanism of action. Use multiple approaches such as pairwise colocalisation tests (using package coloc), statistical fine mapping (using package susieR) and Mandelian Randomisation (using package MandelianRandomization or TwoSampleMR). Take into account linkage disequilibrium (obtain LD matrix with LDlinkR). Use all available datasets, including T1 GWAS, cis-pQTLs, sQTLs and eQTLs. \par
\noindent \textbf{Data Description} \par
There are four datasets available: GWAS summary statistics for myocardial T1, cis\_pQTLs from proteomics GWAS of the same European cohort patients as the myocardial T1 data, sQTLs and eQTLs of genotype-tissue expression data \par

\vspace{1em}

\noindent \textbf{Discovery 5} \par
\noindent \textbf{Research Objective} \par
Identify the most validated Type 2 Diabetes protective mechanisms by ranking SNP-gene pairs based on: (1) multiple independent QTL validations (ATAC, cell eQTL, GTEx eQTL, blood pQTL), (2) directional concordance between GWAS and QTL effect sizes, (3) experimental evidence (ReMap ChIP-seq support), and (4) biological relevance (gene expression levels, posterior probability of causality). For each top mechanism, generate a complete mechanistic report showing: SNP allele —> TF binding
change (with delta scores and cell-type expression) —> chromatin accessibility effects (ATAC betas/ORs
per cell type) —> gene/protein expression changes (with all QTL betas and p-values) —> biological function —> hypothesized link to T2D pathophysiology. Prioritize mechanisms where protective alleles consistently decrease transcript or protein levels across multiple QTL sources, and provide publication-quality visualizations with all statistical evidence clearly labeled. \par
\noindent \textbf{Data Description} \par
You are given two master files, MASTER\_SNP\_GENE\_LEVEL\_FINAL.csv with 12K SNP-gene pairs including ATAC, eQTL, GTEx, and blood pQTL data, and MASTER\_ALL\_CREDIBLE\_SETS\_WITH\_QTLs\_CORRECTED.csv with 68K SNP-TF disruptions including motifBreakR predictions and ReMap ChIP-seq validation, and a comprehensive DATA\_DICTIONARY.csv. \par

\vspace{1em}

\noindent \textbf{Discovery 6} \par
\noindent \textbf{Research Objective} \par
This is a proteomic dataset of mini pools of 10 neurons from Alzheimer's disease cases. Investigate proteome differences between cells with positive and negative tau status. You can also use MC1 (i.e. misfolded tau) quantification to stratify tau positive cells or the precomputed pseudotime to order samples.
Your task is to propose mechanisms contributing to the accumulation of tau and the link with cellular dysfunction, death, or survival. Propose a sequence of events leading to tau accumulation. \par
\noindent \textbf{Data Description} \par
In the obs dataframe, you will find some metadata about samples (age at death, MC1 score, pseudotime), and the tau status (positive/negative). Use robust statistical approaches accounting for the possible cofactors. Data are already log2 transformed. \par

\vspace{1em}

\noindent \textbf{Discovery 7} \par
\noindent \textbf{Research Objective} \par
Investigate differences in transcriptional regulation, transcriptional entropy, transcriptional noise during ageing comparing subclass ‘008 L2/3 IT ENT Glut’ vs ‘007 L2/3 IT CTX Glut’ and ‘003 L5/6 IT TPE-ENT Glut’ vs ‘005 L5 IT CTX Glut’, that could explain a higher propensity to accumulate proteins. Propose mechanisms of vulnerability of ENT and mechanisms of resilience of CTX neurons. \par
\noindent \textbf{Data Description} \par
This is a single nuclei dataset from the ageing mouse brain. In the obs dataframe, you will find some metadata about samples (age, sex), manual cell-type annotation (celltype) and the cell-type annotation using the Allen Brain Institue MapMyCells algorithm (class\_name, subclass\_name, supertype\_name) as well as the cell-typing confidence (\_bootstrapping\_probability). \par

\clearpage
\phantomsection
\subsection*{Supplementary Information 2: Instructions for Evaluations from Academic Groups}
\label{sec:supp_eval_instructions}

\subsubsection*{Comparing with prior research}
\begin{enumerate}
    \item Consider a claim you previously made similar in style and scope to the one from which Kosmos reproduced the discovery. Describe the specific claim that you previously made.
    
    \item How long did it take to produce that claim after you had collected the data? Please answer in terms of elapsed calendar time (days/weeks/months/years).
\end{enumerate}

\subsubsection*{Please review the following Kosmos report. Note that you can click any hyperlinked trajectory to review them.}

\begin{enumerate}
    \setcounter{enumi}{2} 
    \item Please list out the number of all findings in this report that you found to be valuable or significant.
    
    \item Assume your lab received the data on January 1. Under your usual workflow without AI tools, how long would it typically take your lab to generate findings like those in the report, including time spent reading the literature, coming up with ideas, performing the analyses? Please answer in terms of elapsed calendar time in terms of days/weeks/months/years.
    
    \item How many full-time employees would you expect to be working on the project during that period? For example, 1, 1.25, 1.7 FTE.
    
    \item Assess the depth of those findings that you chose, in terms of whether they required deep multi-step reasoning or more shallow ideation:
    \begin{itemize}
        \item Shallow ideation based on surface-level observations
        \item Light reasoning with a straightforward connection of ideas/observations
        \item Moderate depth of reasoning involving several connected steps
        \item Deep multi-step reasoning requiring synthesis across multiple ideas
    \end{itemize}
    
    \item Assess the quality of those findings that you chose in terms of novelty to the field:
    \begin{itemize}
        \item Little novel insight, mostly restatement of known information
        \item Moderate novelty; extends existing knowledge in meaningful ways
        \item Largely novel with minor connections to existing knowledge
        \item Completely novel insight with little precedent
    \end{itemize}
\end{enumerate}

\clearpage
\phantomsection

\subsection*{Supplementary Information 3: Example Instructions for Expert Evaluations (Discovery 1)}
\label{sec:supp_eval_overview}

Thank you for evaluating our scientific findings! We are developing an AI scientist system that can generate novel discoveries. In this project, we performed detailed analyses on the following metabolomics dataset to investigate the specific metabolic adaptations that may enable Preoptic area (POA) Kappa Opioid Receptor (KOR+)-induced cerebroprotection under torpor-like hypothermia and hypometabolism.

The output from the system is a brief report summarizing the main finding, with quantitative details and figures from the analyses. This report contains both results from direct analyses as well as scientific interpretations of the data. We have chosen a random sample of these statements (both based on data analysis and interpretations) for your evaluation:

\begin{enumerate}
    \item Read through the PDF of the discovery group. This report will provide context for the statements that you will be evaluating. The figures are there for your reference, but they will not be evaluated. Do not look through the hyperlinked citations or use those as evidence.
    
    \item Familiarize yourself with the dataset provided. All analyses and discoveries in this discovery group were derived from this dataset.
    
    \item Each discovery group will have a sample of statements to be evaluated. These claims are either derived from data analysis on the dataset, literature review, or by synthesizing ideas across claims/conclusions. Conduct your evaluation based on the type of claim it is. We tried our best to make each statement individually evaluable, but if something is confusing, feel free to refer back to the discovery group PDF.
\end{enumerate}

\subsubsection*{Dataset Information}
This dataset comprises untargeted metabolomic screens of mouse brain tissue, including a polar fraction\\(\texttt{Brain\_Polar\_C\_ST\_NT\_245\_2024.csv}) and a lipid fraction (\texttt{Brain\_Lipids\_C\_ST\_NT.csv}). It contains LC/MS peak intensity measurements for three experimental groups: control (N = 6), hypothermia (N = 6), and normothermia (N = 5). The hypothermia group consists of mice with chemogenetic activation of POA KOR+ neurons that were allowed to undergo a full drop in body temperature, producing a torpor-like hypothermic state. The normothermia group received the same neuronal activation but was housed at an elevated ambient temperature to clamp core temperature near normal, thereby isolating the effects of neural activation without hypothermia and indicating whether protection depends on temperature. The control group received a control treatment without POA KOR+ activation and serves as the baseline for comparison with both the hypothermia and normothermia groups.

\subsubsection*{Evaluation Rubrics}

\textbf{Data Analysis Rubric:} Attempt to replicate this claim using your own data analysis. Did your findings support or refute the claim? The exact numbers and results do not need to match exactly, but your analysis should address the central idea behind the claim. Use your best judgement to determine if the claim is supported, based on the context of the discovery. \textbf{IMPORTANT: Do all data analysis in PYTHON.}
\begin{itemize}
    \item \textbf{Answer:} [SUPPORTED, REFUTED, UNSURE]
    \begin{itemize}
        \item SUPPORTED: My data analysis is able to reproduce this claim.
        \item REFUTED: My data analysis refutes this claim.
        \item UNSURE: I do not know how to do this analysis, the results are unclear/insignificant, or the claim is ambiguous or missing key information.
    \end{itemize}
    \item \textbf{Explanation:} [Free response, backed by evidence.]
    \item \textbf{Your data analysis:} [Jupyter notebook]
\end{itemize}

\textbf{Literature Review Rubric:} Analyze this claim and attempt to verify it in the scientific literature. Does the literature support or refute this claim? If any part of the claim is obviously wrong, it should be marked as refuted. If some parts are supported and others are ambiguous/inconclusive, you may mark it as supported but please use your best judgement.
\begin{itemize}
    \item \textbf{Answer:} [SUPPORTED, REFUTED, UNSURE]
    \begin{itemize}
        \item SUPPORTED: This claim is supported by the literature.
        \item REFUTED: The claim is NOT supported by the literature.
        \item UNSURE: The literature is completely inconclusive, the claim is ambiguous, or unable to be evaluated.
    \end{itemize}
    \item \textbf{Explanation:} [Free response, citing specific papers.]
\end{itemize}

\textbf{Interpretation Rubric:} Analyze the claim and determine if the claim is a scientifically accurate interpretation based only on the report context. This question is only evaluating if the claim’s interpretation of the results is accurate, so you should not conduct any data analysis. A summary of relevant context is given for your reference, which you should assume is correct.
\begin{itemize}
    \item \textbf{Answer:} [SUPPORTED, REFUTED, UNSURE]
    \begin{itemize}
        \item SUPPORTED: This claim is a logically reasonable inference given the report and context.
        \item REFUTED: This claim is NOT a logically reasonable inference given the report and context.
        \item UNSURE: The claim is ambiguous, missing key information, or unable to be evaluated.
    \end{itemize}
    \item \textbf{Explanation:} [Free response, based on the report and context.]
\end{itemize}

\newpage
\subsection*{Supplementary Information 4: Examples of Kosmos evaluations across statement types}
\label{sec:supp_eval_answers}

\subsubsection*{Data Analysis Statement Evaluation Examples}

\vspace{0.5em}
\noindent\textbf{\large Supported}

\vspace{0.5em}
\noindent\textbf{Statement:} When controlling for D‑xylitol, the correlation between betaine and carnosine lost significance and dropped from r=0.910 to r=0.744 (p=0.149).

\vspace{0.5em}
\noindent\textbf{Evaluation:} \textit{I was able to reproduce this claim exactly, including the correlation coefficients and p-values.}

\vspace{0.5em}
\noindent\textbf{\large Refuted}

\vspace{0.5em}
\noindent\textbf{Statement:} In normothermia, PUFA-PC proportion was very strongly and significantly negatively correlated with carnosine.

\vspace{0.5em}
\noindent\textbf{Evaluation:} \textit{It appears that the agent somehow mixed up values for some of the mice, leading the correlation to look a lot stronger than it is. In my analysis, there is a bit of a negative correlation, but it is not significant (p\textasciitilde{}0.3). Overall, the incorrect values negate the claim.}
\vspace{1em}

\subsubsection*{Literature Review Statement Evaluation Examples}

\vspace{0.5em}
\noindent\textbf{\large Supported}

\vspace{0.5em}
\noindent\textbf{Statement:} ASPN-mimic peptides attenuate SMAD2/3 phosphorylation.

\vspace{0.5em}
\noindent\textbf{Evaluation:} \textit{This has been demonstrated across multiple experimental models in both breast cancer and cardiac tissue contexts.  10.1371/journal.pmed.1001871, doi.org/10.1371/journal.pmed.1001871.}

\vspace{0.5em}
\noindent\textbf{\large Refuted}

\vspace{0.5em}
\noindent\textbf{Statement:} Carnosine has specialized roles at the bilayer as a transition-metal chelator and scavenger of secondary non‑aldehyde reactive species arising in arachidonic acid peroxidation cascades.

\vspace{0.5em}
\noindent\textbf{Evaluation:} \textit{It's true that carnosine chelates transition metals. However, it does not seem fair to say that carnosine has specialized roles at the bilayer. First, carnosine is water soluble and its activity for chelating transition metal ions to avoid PUFA (e.g. ARA) peroxidation is likely more about preventing these metals from getting into the membrane. It may be located near the membrane in some cases, though. It can also prevent soluble macromolecule peroxidation (e.g. of proteins), which argues against a specialized role. Additionally, there does not seem to be much support for carnosine scavenging of NON-aldehyde reactive species. It seems key for scavenging aldehydes in particular. Even if I'm overinterpreting the word ``specialized'' here, there does not seem to be much evidence for carnosine scavenging non-aldehydes, so this claim is still not supported.}
\vspace{1em}

\subsubsection*{Interpretation Statement Evaluation Examples}

\vspace{0.5em}
\noindent\textbf{\large Supported}

\vspace{0.5em}
\noindent\textbf{Given context:} Hypothermia uniquely elevated betaine and carnosine and made them strongly coupled, while classic TonEBP osmolytes (myo-inositol, taurine, creatine) did not change.

\vspace{0.5em}
\noindent\textbf{Statement:} The strict state-specificity of this coupling in a design that controls for chemogenetic activation indicates a temperature-dependent metabolic program, rather than an effect of neuronal activation alone.

\vspace{0.5em}
\noindent\textbf{Evaluation:} \textit{Given that both hypothermic and normothermic experimental groups received KOR+ activation, we can resonably conclude that any differences between these groups in metabolites are not due to chemogenetic activation. Given that the key difference is temperature, it is therefore reasonable to say that this coupling indicates temperature-dependent metabolism.}

\vspace{0.5em}
\noindent\textbf{\large Refuted}

\vspace{0.5em}
\noindent\textbf{Given context:} Hypothermia indeed shifted PC composition toward PUFAs relative to normothermia, consistent with a strategic adaptation to maintain membrane fluidity at low temperature. In a targeted comparison, the PUFA-PC proportion was significantly higher in hypothermia than normothermia, with hypothermia showing a mean PUFA proportion of 0.3100 $\pm$ 0.0041 versus 0.2999 $\pm$ 0.0093 in normothermia; the control group was intermediate (0.3063 $\pm$ 0.0063) and not different from either treatment group.

\vspace{0.5em}
\noindent\textbf{Statement:} This modest yet coherent shift implicates compositional remodeling---as opposed to wholesale lipid turnover---as a thermoadaptive mechanism that could help preserve membrane function and provide precursors for protective lipid mediators during the hypometabolic state.

\vspace{0.5em}
\noindent\textbf{Evaluation:} \textit{Despite the caveat of ``modest yet coherent'' in the interpretation, it's unclear to me without further statistics whether this difference is meaningful. The control group PUFA-PC proportion being intermediate additionally argues against this being a mechanistically meaningful change. There are many components to the lipid bilayer and this change may not be very large relative to other changes, some of which could counteract this change. And it may not be large enough to actually preserve membrane function at low temperatures. While PUFAs do maintain membrane fluidity at low temperature and that \textit{could} preserve membrane function, I don't think the context here is sufficient to support this interpretation. Additionally, not all PUFAs equally contribute to supplying protective lipid mediators, so without specific stratified analysis of different PUFA lipid types, it's not possible to evaluate that particular part of this claim. However, it's possible that, if these protective mediator species (ARA/DHA/EPA) are increasing, then this interpretation could be supported. We just don't have enough information.}
\vspace{1em}

\end{document}